\documentclass[10pt,twocolumn,letterpaper]{article}
\usepackage[T1]{fontenc}
\usepackage[pagenumbers]{cvpr}
\usepackage{multirow}
\usepackage{xspace}
\usepackage{microtype}
\usepackage{placeins}

\newcommand{\ours}{MoRAE\xspace}
\newcommand{\pmc}[1]{{\scriptsize$\pm$#1}}

\DeclareMathOperator{\sg}{sg}
\DeclareMathOperator{\KL}{KL}
\usepackage{colortbl}
\newcommand{\defaultrow}{\rowcolor{gray!14}}

\definecolor{linkblue}{rgb}{0.21,0.49,0.74}
\usepackage[pagebackref,hypertexnames=false,breaklinks,colorlinks,allcolors=linkblue]{hyperref}
\usepackage[capitalize]{cleveref}
\hypersetup{
  pdftitle={MoRAE: Representation Autoencoders for Human Motion Generation},
  pdfauthor={Yifei Zhu; Dafei Qin; Zijiao Zeng; Mingyi Shi; Yangyang Cai; Taku Komura; Yoshifumi Kitamura}
}

\title{MoRAE: Representation Autoencoders for Human Motion Generation}
\author{
Yifei Zhu\textsuperscript{1,*} \quad Dafei Qin\textsuperscript{2} \quad Zijiao Zeng\textsuperscript{2} \quad Mingyi Shi\textsuperscript{3}\\
Yangyang Cai\textsuperscript{1} \quad Taku Komura\textsuperscript{3,*} \quad Yoshifumi Kitamura\textsuperscript{1,*}\\[0.3em]
{\normalsize \textsuperscript{1}Tohoku University, Sendai, Japan \quad \textsuperscript{2}Tencent \quad \textsuperscript{3}The University of Hong Kong}\\[0.3em]
{\small \textsuperscript{*}Corresponding authors: \texttt{\href{mailto:yifeizhu@tencent.com}{yifeizhu@tencent.com}\quad
\href{mailto:taku@cs.hku.hk}{taku@cs.hku.hk}\quad
\href{mailto:kitamura@riec.tohoku.ac.jp}{kitamura@riec.tohoku.ac.jp}}}
}

\renewcommand{\dbltopfraction}{0.98}
\renewcommand{\textfraction}{0.02}

\begin{document}
\maketitle
% ===== sec/0_abstract.tex =====
\begin{abstract}
Which latent space should a text-to-motion generator model?
We show that a self-supervised motion representation, once compressed, is a better generative space than latents learned by reconstruction.
Our approach, \ours, is simple: we pre-train a Motion-JEPA encoder that predicts hidden joint trajectories from the rest of the body, and compress its frozen features into a compact latent for a standard flow transformer.
Two observations drive this design.
First, prediction yields features that capture action semantics and cross-joint structure better than reconstruction.
Second, these features are ill-conditioned, and a standard flow transformer generates poorly in them directly. Compressing them into a compact latent makes generation reliable and places cross-joint structure on the directions the flow models most accurately.
With a 58M DiT and the standard CLIP text condition, \ours achieves state-of-the-art FID on HumanML3D and KIT-ML while maintaining retrieval-based text consistency and diversity close to real motion.
\end{abstract}
% ===== end sec/0_abstract.tex =====
% ===== sec/1_intro.tex =====
\begin{figure*}[!t]
  \centering
  \includegraphics[width=\textwidth]{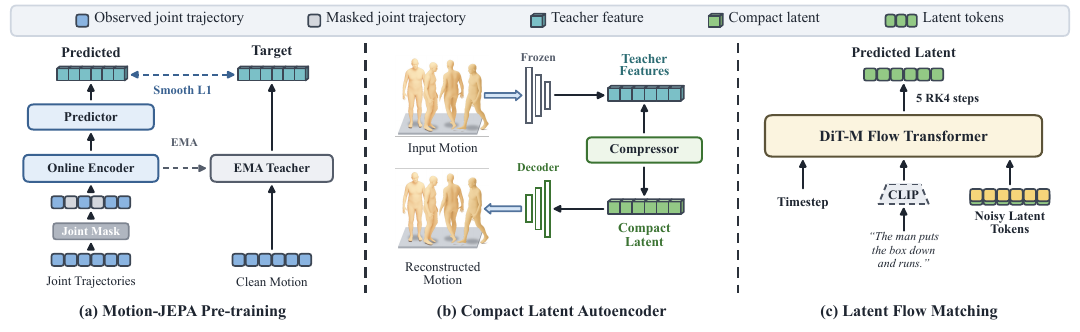}
  \caption{\textbf{Overview of \ours.} (a) \textbf{Motion-JEPA pre-training:} given a clean motion $x$, we mask whole joint trajectories across all channels and frames to form $\tilde{x}$ (masked joint trajectories in gray); an online encoder $f_\theta$ and predictor $p_\phi$ infer the target frame representations $\mathrm{LN}(f_\xi(x))$ of an EMA teacher $f_\xi$ under a smooth $\ell_1$ loss (\cref{eq:jepa}). (b) \textbf{Compact latent autoencoder:} with $f_\xi$ frozen, a single attention layer $E_\psi$ compresses the 768-D teacher features $H\in\mathbb{R}^{T\times 768}$ into a compact 32-D latent $z\in\mathbb{R}^{T\times 32}$, and a lightweight motion decoder $D_\omega$ reconstructs the motion $\hat{x}$ directly in one stage (\cref{eq:tok}). (c) \textbf{Latent flow matching:} conditioned on timestep $t$ and a CLIP text embedding, a 58M DiT-M flow transformer maps noisy latent tokens $z_t=(1-t)\epsilon+tz$ ($\epsilon\sim\mathcal{N}(0,I)$) to the predicted latent $\hat{z}\in\mathbb{R}^{T\times 32}$ in five RK4 steps, which $D_\omega$ decodes into 3D motion (\cref{eq:fm}).}
  \label{fig:overview}
\end{figure*}

\section{Introduction}
\label{sec:intro}
Latent generative models are now the standard route from text to human motion~\cite{chen2023mld,zhang2023t2mgpt,guo2024momask}.
The generator models a latent space learned beforehand, and that choice largely sets the distribution it has to match.
Motion latents are typically learned by reconstruction~\cite{chen2023mld,meng2024rethinking}: an autoencoder recovers its input, with no incentive to organize motion by how the body coordinates its joints.

In images, representation autoencoders take another route and generate in the feature space of a self-supervised encoder~\cite{zheng2025rae,gao2025fae}.
This raises a question for motion: can a self-supervised motion representation serve as a better generative space?
Two conditions must hold: the representation must capture rich motion structure, and a generator must be able to model it accurately.

\begin{figure*}[!t]
  \centering
  \includegraphics[width=\textwidth,trim=0bp 4.4bp 0bp 3.6bp,clip]{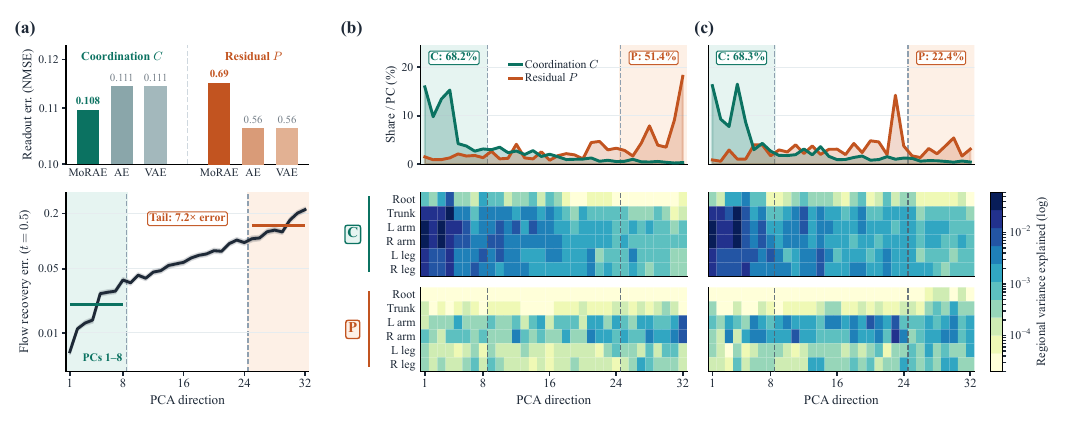}
  \caption{\textbf{Cross-joint readout, flow recovery bias, and spectral organization.} (a) Top: MLP readout error (NMSE) of cross-joint coordination $C(x)$ and joint-specific residual $P(x)=x-C(x)$ from 32-D latents (\cref{tab:rep}); \ours-32 achieves the lowest $C$ error while dropping unpredictable $P$, whereas AE-32 and VAE-32 spend capacity memorizing $P$. Bottom: normalized DiT-M recovery error across the 32 PCA directions ($t{=}0.5$, $w{=}1$), where the trailing quartile (PCs 25--32) has $7.2\times$ the error of the leading quartile (PCs 1--8). (b, c) Directional readout share (top) and regional variance explained (bottom) across the 32 PCA directions for (b) \ours-32 ($\rho(\lambda,L_C/L_P){=}0.918$) and (c) VAE-32 ($\rho{=}0.815$): both models place $C$ in the leading eight directions (68.2\% and 68.3\%), whereas \ours-32 pushes $P$ into the low-variance tail (51.4\% vs.\ 22.4\% for VAE-32).}
  \label{fig:readout}
\end{figure*}

We find that both can be met.
(i) Predicting hidden joints from the rest of the body yields features that make action semantics and cross-joint structure easier to read out than a reconstruction objective does.
We train a Motion-JEPA encoder~\cite{assran2023ijepa} that hides whole joint trajectories and predicts the clean motion representation from the remaining joints; its frozen features recognize actions better than reconstruction-trained encoders of the same architecture and expose the structure shared across joints (\cref{fig:readout}).
(ii) The features are difficult to generate directly with a standard flow transformer.
Like raw motion features, they are numerically rank deficient, and a flow trained on them generates poorly (\cref{fig:spectral}a).
Compression substantially improves generation under the same DiT training and sampling budget.

Driven by these observations, we present \ours, a motion representation autoencoder (\cref{fig:overview}).
\ours compresses frozen Motion-JEPA features into a compact latent with a single attention layer, generates the latent with a standard flow transformer, and decodes it directly to motion with a lightweight decoder.
Why does this compressed latent work?
First, compression removes the rank-deficient tail and leaves a well-conditioned latent that a flow transformer can model; stretching that spectrum while holding width and information fixed is enough to degrade generation.
Second, once the latent is well conditioned, the scalar condition number stops explaining quality; what matters is which motion structure sits on which directions.
At matched width, \ours outperforms reconstruction and other masking schemes because cross-joint coordination lies on the dominant directions, which are the directions a flow recovers most accurately: removing that variance hierarchy by whitening hurts generation, and so does holding the full covariance spectrum fixed while assigning the large variances to directions with weaker cross-joint structure (\cref{sec:recovery}).

\ours focuses on learning an effective motion representation for high-fidelity generation.
With a 58M DiT and standard CLIP sentence conditioning, \ours achieves state-of-the-art FID on HumanML3D (0.0253) and KIT-ML (0.1214).
These results highlight motion representation as a design choice complementary to the generator and text-conditioning mechanism.

% ===== end sec/1_intro.tex =====
% ===== sec/2_related.tex =====
\section{Related Work}
\label{sec:related}
\paragraph{Text-to-motion generation.}
Diffusion and flow models generate motion directly in raw motion space~\cite{tevet2023mdm,zhang2022motiondiffuse,zhang2023remodiffuse,li2026motionhiflow} or in a continuous latent space~\cite{chen2023mld,motionlcm,meng2024rethinking,hong2025salad}.
Discrete approaches learn motion codebooks and generate tokens autoregressively or by masked modeling~\cite{zhang2023t2mgpt,guo2024momask,pinyoanuntapong2024mmm,bamm,yuan2024mogents,guo2025snapmogen}, while Light-T2M studies compact generators~\cite{zeng2025lightt2m}.
In all of these models, the latent is typically learned by reconstruction; recent work further shapes it with text semantics~\cite{he2026molingo} or codebook geometry~\cite{fang2026mogeflow}.
\ours learns its latent from a self-supervised motion representation.

\paragraph{Representation autoencoders.}
Latent diffusion typically trains a generator in the latent space of an autoencoder learned by reconstruction~\cite{rombach2022ldm}.
In images, one line of work aligns diffusion features~\cite{yu2024repa} or autoencoder latents~\cite{yao2025lightningdit} with a pretrained representation.
RAE~\cite{zheng2025rae,singh2026improvedbaselinesrepresentationautoencoders} goes further and generates directly in the feature space of a frozen self-supervised encoder, while FAE~\cite{gao2025fae} compresses those features with a single attention layer and decodes in two stages (first back to high-dimensional features, then to pixels).
Motion does not transfer either recipe unchanged: unlike visual patch features, self-supervised motion features are numerically rank deficient, so generating directly in the frozen feature space fails (\cref{sec:kappa}).
Our use of compression also differs from FAE: \ours decodes the compressed latent to motion in a single stage, outperforming two-stage feature decoding (\cref{sec:supp_geometry}), and shows that compression works not merely by lowering dimensionality or condition number, but by preserving a variance hierarchy that places cross-joint coordination on the dominant directions best recovered by the flow (\cref{sec:recovery}).

\paragraph{Self-supervised motion representations.}
Masked autoencoders reconstruct hidden inputs~\cite{he2022mae}, whereas joint-embedding methods predict the representations of hidden inputs~\cite{assran2023ijepa,bardes2024vjepa,baevski2022data2vec}.
On skeletons, prior methods reconstruct masked joints~\cite{yan2023skeletonmae}, predict temporal motion~\cite{mao2023mamp}, or lift 2D poses to 3D~\cite{zhu2023motionbert} for recognition.
Closest to our encoder, S-JEPA~\cite{abdelfattah2024sjepa} applies spatio-temporal masking to 3D joint coordinates for action classification.
Generative motion features, however, pack each frame into one token in which every joint is written several times through its position, rotation, velocity, and contact.
Under spatio-temporal or partial-channel masking, the encoder can copy missing values from adjacent frames or from other channels of the same joint, so it never has to represent how the rest of the body constrains that joint (\cref{sec:ablation}).
\ours instead masks each joint's entire trajectory across all channels and frames, forcing the encoder to capture whole-body cross-joint coordination.
% ===== end sec/2_related.tex =====
% ===== sec/3_method.tex =====
\section{Method}
\label{sec:method}
\ours is a representation autoencoder for motion (\cref{fig:overview}).
It has three parts, trained in sequence: a Motion-JEPA encoder that learns a motion representation by predicting hidden joints, a compressor and decoder that turn the frozen representation into a compact latent and back into motion, and a flow transformer that generates the latent from text.

\paragraph{Motion and joints.}
HumanML3D describes each frame by $D{=}263$ features~\cite{guo2022humanml3d}: root velocity and height, joint positions, rotations, velocities, and foot contacts.
We assign every channel to a joint, partitioning the features into $J{=}22$ joint blocks $g_1,\dots,g_J$; the root block holds the root channels, and each foot block holds that foot's contact channel.
KIT-ML uses the same layout with $J{=}21$ and $D{=}251$.

\subsection{Motion-JEPA}
\label{sec:jepa}
\paragraph{Masking.}
For a motion $x\in\mathbb{R}^{T\times D}$, we sample $\lfloor 0.3J\rceil$ joints and replace all of their channels in every frame with learned mask values, calling the result $\tilde{x}$.
The joint is the natural unit to mask: position, rotation, and velocity are three views of the same joint, so masking only some channels or some frames leaves a local shortcut.
Hiding the whole trajectory forces the encoder to infer that joint from the rest of the body (\cref{sec:ablation}).

\paragraph{Prediction.}
Following I-JEPA~\cite{assran2023ijepa}, an encoder $f_\theta$ reads $\tilde{x}$ and a predictor $p_\phi$ predicts the representation that an EMA teacher $f_{\xi}$ computes from the clean motion:
\begin{equation}
  \mathcal{L}_{\text{JEPA}} = \frac{1}{T}\sum_{t=1}^{T} \ell_1^{\text{smooth}}\!\Big( p_\phi\big(f_\theta(\tilde{x})\big)_t,\ \sg\big[\mathrm{LN}(f_{\xi}(x))_t\big] \Big).
  \label{eq:jepa}
\end{equation}
The encoder is a Transformer with 8 layers of width 768 and one token per frame, the predictor has 6 layers of width 384, and the teacher momentum rises from 0.996 to 1.
After pre-training, the teacher provides frame features $H=\mathrm{LN}\big(\sum_{l=5}^{8} h^{(l)}\big)\in\mathbb{R}^{T\times 768}$ from its last four layers.

\subsection{Compact latent}
\label{sec:code}
Because the 768-D features are numerically rank deficient and difficult to generate directly (\cref{sec:kappa}), we compress them into a low-dimensional latent.
With the encoder frozen, a single attention layer $E_\psi$~\cite{gao2025fae} maps $H$ to a diagonal Gaussian posterior over a 32-D latent per frame, and a motion decoder $D_\omega$ (4 Transformer layers, width 512) maps the latent back to all motion channels:
\begin{equation}
  \mathcal{L}_{\text{AE}} = \big\| D_\omega(z) - x \big\|_2^2 + \beta\, \KL\big(q_\psi(z\,|\,H)\,\|\,\mathcal{N}(0,I)\big),
  \label{eq:tok}
\end{equation}
with $z\sim q_\psi(z|H)$ and $\beta{=}10^{-4}$.
The posterior mean serves as the latent for generation.

\subsection{Latent flow matching}
\label{sec:flow}
We generate the latent with a standard flow transformer.
Each latent channel is standardized with training statistics.
The generator, DiT-M, is a Transformer with 12 layers of width 512 and 8 heads (58M parameters) over one token per frame; the timestep and the CLIP ViT-B/32 sentence embedding~\cite{radford2021clip} condition every layer through adaptive layer normalization~\cite{peebles2023dit}.
With the linear path $z_t=(1-t)\,\epsilon + t\,z$, $\epsilon\sim\mathcal{N}(0,I)$, the network regresses the velocity~\cite{lipman2023flowmatching,ma2024sit}:
\begin{equation}
  \mathcal{L}_{\text{FM}} = \mathbb{E}_{t,\epsilon,z,c}\,\big\| v_\vartheta(z_t, t, c) - (z-\epsilon) \big\|_2^2 .
  \label{eq:fm}
\end{equation}
Text is dropped with probability 0.1 for classifier-free guidance~\cite{cfg}.
At inference, we integrate the guided velocity from noise with five RK4 steps (20 guided evaluations) and decode with $D_\omega$; only CLIP, the DiT, and the decoder are used.
% ===== end sec/3_method.tex =====
% ===== sec/5_analysis.tex =====
\section{Experiments}
\label{sec:analysis}
\label{sec:exp}
\begin{figure*}[!t]
  \centering
  \includegraphics[width=\textwidth]{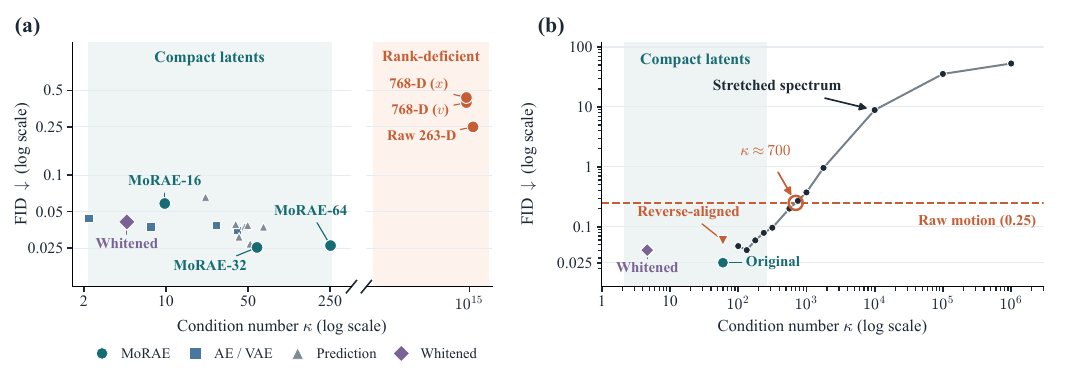}
  \caption{\textbf{Spectral conditioning and coordinate interventions.} (a) FID versus condition number $\kappa$ across the 18 generative spaces (\cref{tab:supp_all,tab:supp_kappa_all}): rank-deficient uncompressed spaces fail, whereas compact latents ($\kappa\in[2.2,252]$) generate well, with \ours-32 (0.0253) and \ours-64 (0.0262) achieving the lowest FID. (b) On \ours-32 ($d{=}32$), stretching $\kappa$ from $10^2$ to $10^6$ crosses the raw-motion baseline near $\kappa\approx700$, while whitening ($\kappa{=}4.65$) and reverse-aligned fixed-spectrum coordinates ($\kappa{=}59.70$) degrade FID relative to the original coordinates (\cref{tab:white,tab:iso,tab:supp_kappa_ctrl}).}
  \label{fig:spectral}
\end{figure*}
\paragraph{Setup.}
\label{sec:setup}
We use the standard splits and evaluation protocol of HumanML3D~\cite{guo2022humanml3d} and KIT-ML~\cite{kit}, reporting FID, R-Precision at ranks 1--3, multimodal distance (MMDist), and Diversity, each averaged over 20 sampling repetitions with a 95\% confidence interval, except the coordinate--metric control below.
Guidance, the number of integration steps, and generator checkpoints are selected on the validation split; every reported number is a test-split evaluation.
On each dataset, we train the encoder (600 epochs), the compressor and decoder (400 epochs), and the generator (500 epochs).
Unless noted, latents are 32-D and generated by DiT-M with five RK4 steps.
Implementation details are in \cref{sec:supp_impl}, and analyses of the latent use a fixed HumanML3D diagnostic split (\cref{sec:supp_analysis}).

\subsection{Representation}
\label{sec:coord}
% ===== sec/table_rep.tex =====
\begin{table}[t]
\centering\footnotesize
\setlength{\tabcolsep}{3pt}
\begin{tabular}{lccccc}
\toprule
 & Action & \multicolumn{2}{c}{Readout err.\ $\downarrow$} & \multicolumn{2}{c}{Spectral order} \\
\cmidrule(lr){3-4}\cmidrule(lr){5-6}
Latent & acc.\ $\uparrow$ & $C$ & $P$ & $\rho(\lambda,L_P)$ & $\rho(\lambda,L_C/L_P)$ \\
\midrule
AE-32 & 77.8\% & 0.1111 & 0.5563 & $-0.387$ & 0.856 \\
VAE-32 & 78.8\% & 0.1112 & \textbf{0.5562} & $-0.293$ & 0.815 \\
\defaultrow \ours-32 & \textbf{82.1\%} & \textbf{0.1078} & 0.6895 & $\mathbf{-0.694}$ & \textbf{0.918} \\
\bottomrule
\end{tabular}
\caption{\textbf{Representation structure on HumanML3D.} Action linear-probe accuracy, MLP readout error of $C$ and $P$, and spectral rank correlations.}
\label{tab:rep}
\end{table}
% ===== end sec/table_rep.tex =====
\paragraph{Action semantics.}
A generative motion representation should capture high-level action semantics.
We test this with a linear probe on frozen encoder features across 12 HumanML3D action classes (\cref{tab:rep,tab:supp_semantic}).
Motion-JEPA is more linearly separable than reconstruction-trained AE and VAE encoders of the same architecture, and substantially more separable than raw motion (70.9\%).
Predicting masked joints makes action semantics easier to read out than reconstructing the input.

\paragraph{Coordination versus residual detail.}
Why does joint prediction change what the compact latent retains?
We split each pose $x$ into cross-joint coordination $C(x)$, the component a ridge regression predicts from the other 21 joints at the same frame, and the joint-specific residual $P(x)=x-C(x)$; even a five-frame nonlinear probe leaves most of $P(x)$ unexplained (\cref{tab:supp_context}).
Probes on the 32-D latents show that the two pre-training objectives keep different pieces (\cref{fig:readout}a, \cref{tab:rep,tab:supp_readout,tab:supp_budget}).
Because AE and VAE must reconstruct every input channel, they spend latent capacity on the unpredictable residual $P$ and achieve a lower $P$ readout error.
Motion-JEPA only predicts what the rest of the body determines, so \ours prioritizes cross-joint coordination $C$ and achieves the lowest $C$ readout error (with a paired 95\% CI against VAE entirely below zero).
Predictive pre-training thus dedicates the compact latent to whole-body coordination.

\paragraph{Spectral organization.}
Beyond what the latent retains overall, pre-training decides where $C$ and $P$ sit in the spectrum.
We fit PCA in the normalized flow coordinates ($\lambda_1\geq\dots\geq\lambda_{32}$) and measure the fraction of each target explained along each direction (\cref{fig:readout}b,c, \cref{tab:rep,tab:supp_spectral}).
Every model places coordination on its leading directions (the top eight components capture 68\% of $C$ for \ours), but AE and VAE scatter the residual $P$ through high- and mid-variance directions.
\ours instead pushes $P$ into the low-variance tail, placing 51.4\% of $P$ in the last eight components against 22.4\% for VAE, so the coordination-to-residual ratio tracks the eigenvalue order (\cref{tab:rep}).
The \ours spectrum thus establishes a clean variance hierarchy: dominant directions carry whole-body coordination, whereas trailing directions hold joint-specific residuals.

\subsection{Compression}
\label{sec:kappa}
\label{sec:aniso}
% ===== sec/table_width.tex =====
\begin{table}[t]
\centering\small
\setlength{\tabcolsep}{5pt}
\begin{tabular}{lccc}
\toprule
Generative space & $d$ & $\kappa$ & FID $\downarrow$ \\
\midrule
Motion-JEPA features & 768 & rank def. & 0.3970\pmc{.0111} \\
Direct 32-D bottleneck & 32 & 49.68 & 0.0385\pmc{.0018} \\
\ours-16 & 16 & 9.78 & 0.0501\pmc{.0021}$^\dagger$ \\
\defaultrow \ours-32 & 32 & 59.70 & 0.0253\pmc{.0015} \\
\ours-64 & 64 & 252.19 & 0.0262\pmc{.0011} \\
\bottomrule
\end{tabular}
\caption{\textbf{Compression and latent width} on HumanML3D ($^\dagger$12 RK4 steps; default in gray).}
\label{tab:width}
\end{table}
% ===== end sec/table_width.tex =====
\paragraph{Conditioning.}
On the path $z_t=(1-t)\epsilon+tz$, PCA direction $i$ has signal-to-noise ratio $\mathrm{SNR}_i(t)=t^2\lambda_i/(1-t)^2$.
Under a covariance-matched Gaussian reference, its optimal normalized recovery error and the time at which its signal and noise variances are equal are
\begin{equation}
 e_i^*(t)=\frac{1}{1+\mathrm{SNR}_i(t)},\qquad
 t_i^*=\frac{1}{1+\sqrt{\lambda_i}}.
 \label{eq:spectral_recovery}
\end{equation}
Low-variance directions resolve late, and $\kappa=\lambda_1/\lambda_d$ measures how widely the resolution times spread (\cref{sec:supp_spectral_theory}).

\paragraph{Uncompressed features.}
Raw motion and Motion-JEPA features are numerically rank deficient, with eigenvalue ratios near $10^{15}$.
The deficiency spans a long tail: raw motion has 89 directions, and Motion-JEPA features have 312, carrying less than $10^{-3}$ of the leading variance, whereas no compact latent has any.
Because those trailing directions resolve only at the end of the path, five-step flows trained directly on either uncompressed space fail (\cref{fig:spectral}a, \cref{tab:supp_all}), whereas the same DiT training and sampling budget succeeds once the features are compressed.

\paragraph{Latent width.}
Pre-training and generation favor different widths: masked prediction benefits from a wide encoder, whereas flow matching benefits from a compact latent.
Compressing the 768-D encoder features to 32 dimensions achieves the best FID and saturates at 64 dimensions, while a 32-D bottleneck trained directly inside the encoder is substantially worse (\cref{tab:width}).
Keeping a wide predictive encoder and a separate 32-D compressor lets the latent stay informative while remaining easy for the flow to model.

\paragraph{Conditioning at fixed width.}
To separate ill-conditioning from dimensionality and information content, we stretch the spectrum of \ours-32 with invertible linear maps and fold the inverse into the decoder so reconstructions remain identical (\cref{fig:spectral}b, \cref{sec:supp_fixedwidth}).
Under the same DiT-M recipe, FID stays stable through moderate stretching, then rises sharply and reaches the five-step raw-motion baseline (0.25) near $\kappa\approx700$ (\cref{tab:supp_kappa_ctrl}).
Extreme spectral imbalance alone is therefore enough to break flow generation.

Learned compact latents all sit well below that collapse ($\kappa\in[2.2, 252]$).
Inside this well-conditioned range, the scalar condition number no longer tracks quality: \ours-32 and \ours-64 have larger $\kappa$ than the reconstruction and 16-D baselines, yet they achieve the lowest FIDs (\cref{tab:width}).
From here, quality follows how pre-training organizes motion structure across directions (\cref{sec:ablation,sec:recovery}).

\subsection{Pre-training}
\label{sec:ablation}
% ===== sec/table_objective.tex =====
\begin{table}[t]
\centering\small
\setlength{\tabcolsep}{5pt}
\begin{tabular}{lcc}
\toprule
Latent objective & HumanML3D & KIT-ML \\
\midrule
Reconstruction (AE) & 0.0349\pmc{.0021} & 0.1330\pmc{.0045} \\
Reconstruction (VAE) & 0.0383\pmc{.0019} & 0.1443\pmc{.0066} \\
\defaultrow Prediction (\ours) & \textbf{0.0253}\pmc{.0015} & \textbf{0.1214}\pmc{.0037} \\
\bottomrule
\end{tabular}
\caption{\textbf{Pre-training objective} (FID $\downarrow$) with matched 32-D architecture and DiT-M recipe.}
\label{tab:objective}
\end{table}
% ===== end sec/table_objective.tex =====
% ===== sec/table_masking.tex =====
\begin{table}[t]
\centering\small
\setlength{\tabcolsep}{6pt}
\begin{tabular}{lc}
\toprule
Masking unit & FID $\downarrow$ \\
\midrule
Temporal spans & 0.0584\pmc{.0025}$^\dagger$ \\
Scrambled channel groups & 0.0377\pmc{.0019} \\
Joints and temporal spans & 0.0271\pmc{.0016} \\
\defaultrow Whole joint trajectories (\ours) & \textbf{0.0253}\pmc{.0015} \\
\bottomrule
\end{tabular}
\caption{\textbf{Masking unit} on HumanML3D ($^\dagger$12 RK4 steps; default in gray).}
\label{tab:masking}
\end{table}
% ===== end sec/table_masking.tex =====
% ===== sec/table_main.tex =====
\begin{table*}[t]
\centering\small
\setlength{\tabcolsep}{5pt}
\begin{tabular}{lcccccc}
\toprule
 & & \multicolumn{3}{c}{R Precision$\uparrow$} & & \\
\cmidrule(lr){3-5}
Method & FID $\downarrow$ & Top 1 & Top 2 & Top 3 & MMDist $\downarrow$ & Diversity $\to$ \\
\midrule
\multicolumn{7}{l}{\textit{HumanML3D}} \\
Real motion & 0.002\pmc{.000} & 0.511\pmc{.003} & 0.703\pmc{.003} & 0.797\pmc{.002} & 2.974\pmc{.008} & 9.503\pmc{.065} \\
MLD~\cite{chen2023mld} & 0.473\pmc{.013} & 0.481\pmc{.003} & 0.673\pmc{.003} & 0.772\pmc{.002} & 3.196\pmc{.010} & 9.724\pmc{.082} \\
T2M-GPT~\cite{zhang2023t2mgpt} & 0.141\pmc{.005} & 0.492\pmc{.003} & 0.679\pmc{.002} & 0.775\pmc{.002} & 3.121\pmc{.009} & 9.761\pmc{.081} \\
MoMask~\cite{guo2024momask} & 0.045\pmc{.002} & 0.521\pmc{.002} & 0.713\pmc{.002} & 0.807\pmc{.002} & 2.958\pmc{.008} & -- \\
BAMM~\cite{bamm} & 0.055\pmc{.002} & 0.525\pmc{.002} & 0.720\pmc{.003} & 0.814\pmc{.003} & 2.919\pmc{.008} & 9.717\pmc{.089} \\
MoGenTS~\cite{yuan2024mogents} & 0.033\pmc{.001} & 0.529\pmc{.003} & 0.719\pmc{.002} & 0.812\pmc{.002} & 2.867\pmc{.006} & \underline{9.570}\pmc{.077} \\
Light-T2M~\cite{zeng2025lightt2m} & 0.040\pmc{.002} & 0.511\pmc{.003} & 0.699\pmc{.002} & 0.795\pmc{.002} & 3.002\pmc{.008} & -- \\
SALAD~\cite{hong2025salad} & 0.076\pmc{.002} & \underline{0.581}\pmc{.003} & \underline{0.769}\pmc{.003} & \underline{0.857}\pmc{.002} & \underline{2.649}\pmc{.009} & 9.696\pmc{.096} \\
MotionHiFlow~\cite{li2026motionhiflow} & \underline{0.032}\pmc{.002} & 0.563\pmc{.003} & 0.754\pmc{.003} & 0.843\pmc{.003} & 2.691\pmc{.009} & \textbf{9.504}\pmc{.071} \\
MoGeFlow~\cite{fang2026mogeflow} & 0.048\pmc{.003} & \textbf{0.589}\pmc{.002} & \textbf{0.784}\pmc{.002} & \textbf{0.874}\pmc{.002} & \textbf{2.600}\pmc{.008} & 9.914\pmc{.081} \\
\cmidrule(lr){1-7}
\ours (32-D) & \textbf{0.0253}\pmc{.0015} & 0.520\pmc{.002} & 0.709\pmc{.002} & 0.801\pmc{.002} & 2.939\pmc{.005} & 9.649\pmc{.072} \\
\midrule
\multicolumn{7}{l}{\textit{KIT-ML}} \\
Real motion & 0.031\pmc{.004} & 0.424\pmc{.005} & 0.649\pmc{.006} & 0.779\pmc{.006} & 2.788\pmc{.012} & 11.08\pmc{.097} \\
MLD~\cite{chen2023mld} & 0.404\pmc{.027} & 0.390\pmc{.008} & 0.609\pmc{.008} & 0.734\pmc{.007} & 3.204\pmc{.027} & 10.80\pmc{.117} \\
T2M-GPT~\cite{zhang2023t2mgpt} & 0.514\pmc{.029} & 0.416\pmc{.006} & 0.627\pmc{.006} & 0.745\pmc{.006} & 3.007\pmc{.023} & 10.921\pmc{.108} \\
MoMask~\cite{guo2024momask} & 0.204\pmc{.011} & 0.433\pmc{.007} & 0.656\pmc{.005} & 0.781\pmc{.005} & 2.779\pmc{.022} & -- \\
BAMM~\cite{bamm} & 0.183\pmc{.013} & 0.438\pmc{.009} & 0.661\pmc{.009} & 0.788\pmc{.005} & 2.723\pmc{.026} & 11.008\pmc{.094} \\
MoGenTS~\cite{yuan2024mogents} & 0.143\pmc{.004} & 0.445\pmc{.006} & 0.671\pmc{.006} & 0.797\pmc{.005} & 2.711\pmc{.024} & 10.918\pmc{.090} \\
Light-T2M~\cite{zeng2025lightt2m} & 0.161\pmc{.009} & 0.444\pmc{.006} & 0.670\pmc{.007} & 0.794\pmc{.005} & 2.746\pmc{.016} & -- \\
SALAD~\cite{hong2025salad} & 0.296\pmc{.012} & 0.477\pmc{.006} & \underline{0.711}\pmc{.005} & \underline{0.828}\pmc{.005} & 2.585\pmc{.016} & \underline{11.097}\pmc{.095} \\
MotionHiFlow~\cite{li2026motionhiflow} & \underline{0.135}\pmc{.007} & \textbf{0.482}\pmc{.005} & 0.704\pmc{.005} & 0.825\pmc{.005} & \underline{2.552}\pmc{.014} & 10.894\pmc{.117} \\
MoGeFlow~\cite{fang2026mogeflow} & 0.145\pmc{.005} & \underline{0.479}\pmc{.007} & \textbf{0.712}\pmc{.006} & \textbf{0.842}\pmc{.004} & \textbf{2.499}\pmc{.016} & 10.724\pmc{.093} \\
\cmidrule(lr){1-7}
\ours (32-D) & \textbf{0.1214}\pmc{.0037} & 0.478\pmc{.005} & 0.700\pmc{.006} & 0.822\pmc{.004} & \underline{2.552}\pmc{.018} & \textbf{11.075}\pmc{.145} \\
\bottomrule
\end{tabular}
\caption{\textbf{Comparison on HumanML3D and KIT-ML.} Means and 95\% CIs over 20 evaluations. \textbf{Bold}: best; \underline{underlined}: second best; $\to$: closer to real motion is better.}
\label{tab:main}
\end{table*}
% ===== end sec/table_main.tex =====
\paragraph{Prediction versus reconstruction.}
With the encoder architecture and 32-D width matched, prediction outperforms reconstruction on both HumanML3D (lowering FID by 27.5\% over AE and 33.9\% over VAE) and KIT-ML (by 8.7\% over AE and 15.9\% over VAE), even though the reconstruction latents have smaller condition numbers (\cref{tab:objective,tab:supp_kappa_all}).
Raising the VAE KL weight drives $\kappa$ down to $2.2$ and still never brings FID below 0.0338 (\cref{tab:supp_all}).
The mask is not the missing ingredient either: reconstructing raw channels under the same whole-joint mask, as in an MAE, is worse than predicting the latent target (FID 0.0402; \cref{tab:supp_mae}).
Predicting in representation space is what removes the pressure to memorize residuals $P(x)$ that the rest of the body does not determine.

\paragraph{Masking unit.}
Prediction captures whole-body coordination only when the mask removes local shortcuts.
Masking temporal spans leaves neighboring frames to interpolate from, and masking scrambled channel groups leaves other channels of the same joint as shortcuts (\cref{tab:masking}).
Hiding the entire joint trajectory across all channels and frames achieves the best FID, and adding temporal masking on top brings no further gain.
The anatomical joint trajectory is thus the unit that forces the encoder to infer the missing joint from cross-joint coordination $C(x)$.

\subsection{Why spectral organization governs generation}
\label{sec:recovery}
\begin{figure*}[!t]
  \centering
  \includegraphics[width=\textwidth]{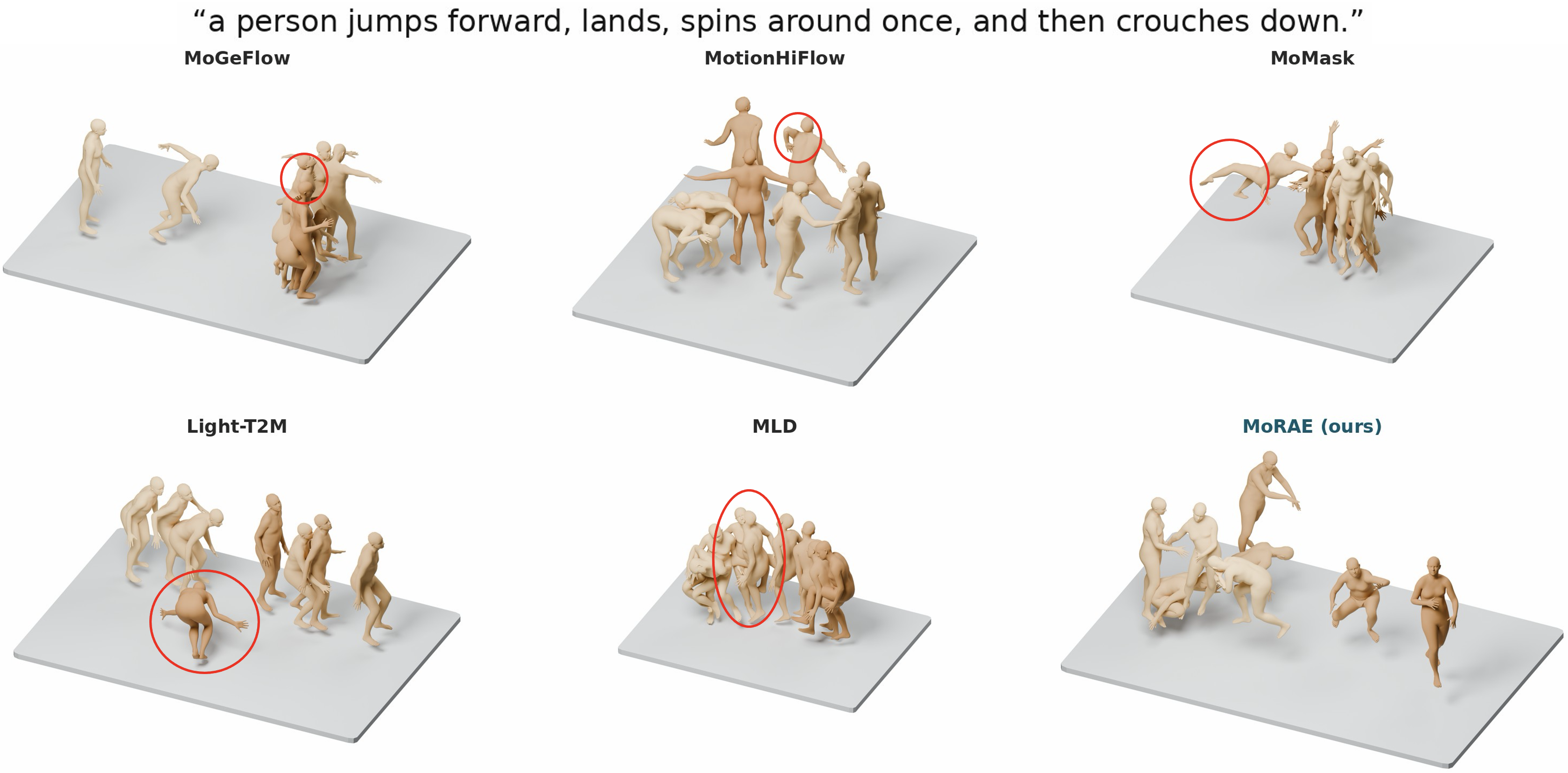}
  \caption{\textbf{Qualitative comparison with prior methods.} For the multi-action prompt ``a person jumps forward, lands, spins around once, and then crouches down,'' every method generates one sample with the same random seed, shown at eight evenly spaced frames. \ours carries a coherent body through the jump, the spin, and the crouch, whereas prior methods produce distorted body configurations (circled).}
  \label{fig:qual}
\end{figure*}
% ===== sec/table_whitening.tex =====
\begin{table}[t]
\centering\footnotesize
\setlength{\tabcolsep}{3pt}
\begin{tabular}{lccccc}
\toprule
 & & & \multicolumn{2}{c}{Decoded err.\ $\downarrow$} & \\
\cmidrule(lr){4-5}
\ours-32 & $\kappa$ & $\rho(\lambda,L_C/L_P)$ & $C$ & $P$ & FID $\downarrow$ \\
\midrule
\defaultrow Original & 59.70 & \textbf{0.918} & \textbf{0.228} & 0.723 & \textbf{0.0253}\pmc{.0015} \\
Whitened & \textbf{4.65} & 0.229 & 0.256 & \textbf{0.675} & 0.0411\pmc{.0023} \\
\bottomrule
\end{tabular}
\caption{\textbf{Effect of whitening \ours-32} on HumanML3D (decoded motion unchanged; \cref{sec:supp_white}).}
\label{tab:white}
\end{table}
% ===== end sec/table_whitening.tex =====
\paragraph{Directional recovery bias of the flow.}
Spectral organization matters because a finite-capacity flow model does not recover every latent direction equally well: it is far more accurate on high-variance directions than on low-variance ones.
By \cref{eq:spectral_recovery}, high-variance directions reach unit SNR early along the flow path, whereas low-variance trailing directions stay buried in noise until $t\to 1$.
We verify this on the trained DiT-M using the one-step target estimate along $z_t=(1-t)\epsilon+tz$:
\begin{equation}
 \hat z=z_t+(1-t)\,v_\vartheta(z_t,t,c).
 \label{eq:recovery}
\end{equation}
After each PCA direction's squared error is normalized by its training variance $\lambda_i$, the lowest-variance quartile (PCs 25--32) has $7.2\times$ the error of the highest-variance quartile (PCs 1--8) at $t=0.5$ and $w{=}1$ (\cref{fig:readout}a, \cref{fig:supp_directional}a, \cref{fig:supp_heatmap}).
The flow generator is therefore inherently most reliable on its dominant PCA directions.

\paragraph{Aligning motion structure with flow recovery.}
This recovery bias explains why the spectral organization of $C$ and $P$ (\cref{sec:coord}) governs generation quality.
Because \ours concentrates cross-joint coordination $C$ in the high-variance leading directions and pushes residuals $P$ into the tail (\cref{tab:rep}), the flow generator spends its high-SNR precision on global body coherence while leaving fine per-joint residuals to the deterministic motion decoder.
By contrast, AE and VAE mix unpredictable residuals $P$ into their higher-variance directions, wasting high-SNR generator capacity on local joint variations.

\paragraph{Flattening the hierarchy via whitening.}
To test whether this variance hierarchy itself, rather than merely the subspace spanned by the 32-D latent, drives the generation gain, we apply the invertible whitening map $z_w=\Lambda^{-1/2}U^\top(z-\mu)$ and fold its inverse into the decoder (\cref{sec:supp_white}).
Whitening leaves the decoded motions unchanged and erases the alignment between variance and cross-joint coordination, raising DiT-M FID by 62\% (\cref{tab:white}).
Decoded error moves with that flattening: error on the local residual $P$ falls, while error on whole-body coordination $C$ rises (\cref{tab:white,fig:supp_directional,fig:supp_heatmap}).
Because the flow-matching objective is computed in the new coordinates, whitening changes both the coordinates in which the flow evolves and the directional weights of the loss; the next control separates those two effects.

\paragraph{Coordinates versus training metric.}
We cross original and whitened coordinates with squared-error losses defined in the original and whitened spaces (\cref{sec:supp_coord_loss,tab:supp_coord_loss}).
On the shared 4,382-clip, 15-seed protocol, the original coordinates with the original-space loss reach FID 0.0244 (corresponding to 0.0253 for this model over 20 seeds on the full test split).
In those original coordinates, switching to the whitened-space loss upweights the low-variance tail and raises FID to 0.0659.
In whitened coordinates, the two losses agree: 0.0407 with the original-space loss and 0.0416 with the whitened-space loss, both well behind 0.0244.
The loss metric matters when the variance hierarchy is present, and flattening the coordinates still hurts after the original-space loss is restored.
Because that cross still alters the eigenvalue spectrum, we next hold the spectrum fixed and change only which directions receive the large variances.

% ===== sec/table_isospectral.tex =====
\begin{table}[t]
\centering\small
\setlength{\tabcolsep}{4pt}
\begin{tabular}{lccc}
\toprule
Spectrum assignment & FID $\downarrow$ & $C$ err.\ $\downarrow$ & $P$ err.\ $\downarrow$ \\
\midrule
\defaultrow $C$-aligned & \textbf{0.0279}\pmc{.0016} & \textbf{0.2184} & \textbf{0.5617} \\
Reverse-aligned & 0.0589\pmc{.0030} & 0.3471 & 0.6515 \\
\bottomrule
\end{tabular}
\caption{\textbf{Alignment at a fixed spectrum} ($\kappa=59.70$) on HumanML3D. $C/P$ errors are relative to the shared autoencoder reconstruction (\cref{sec:supp_iso}).}
\label{tab:iso}
\end{table}
% ===== end sec/table_isospectral.tex =====
\paragraph{Alignment at a fixed spectrum.}
When original PCA direction $i$ has cross-joint loading $L_C(i)$ and assigned variance $\nu_i$, the Gaussian reference recovery error (\cref{eq:spectral_recovery}) for the linear readout of $C$ is
\begin{equation}
 \mathcal{E}_{C,\mathrm{lin}}^*(t;\nu)
 =\sum_{i=1}^d L_C(i)\,e_{i,\nu}^*(t)
 =\sum_{i=1}^d \frac{L_C(i)}{1+t^2\nu_i/(1-t)^2}.
 \label{eq:rearrangement}
\end{equation}
Holding the eigenvalue multiset $\{\nu_i\}$ fixed preserves $\kappa$ and the aggregate unweighted Gaussian recovery error, while changing how recovery accuracy is allocated across the original PCA directions; by the rearrangement inequality, $\mathcal{E}_{C,\mathrm{lin}}^*(t;\nu)$ is minimized when larger variances $\nu_i$ sit on larger loadings $L_C(i)$ and maximized under the reverse pairing (\cref{sec:supp_spectral_theory}).
We test this prediction without changing the spectrum: freezing the \ours-32 compressor and decoder, we reassign the exact same 32 eigenvalues to the original PCA directions in descending ($C$-aligned) or ascending (Reverse-aligned) order of $L_C(i)$ and apply the inverse scaling before the shared decoder so reconstructions match (\cref{tab:iso,sec:supp_iso}).
At this shared spectrum ($\kappa=59.70$), $C$-aligned lowers FID by 52.6\% relative to Reverse-aligned and reduces time-integrated decoded error on both $C$ (by 37.1\%) and $P$ (by 13.8\%; \cref{tab:iso,tab:supp_iso_rec}).
With the entire eigenvalue spectrum held fixed, generation quality follows from pairing the high-variance directions that the flow recovers best with cross-joint coordination; consistently, a geometric spectrum at $\kappa=10^2$ with a random rotation that misaligns coordination and high variance also degrades generation (\cref{tab:supp_kappa_ctrl}).
% ===== end sec/5_analysis.tex =====
% ===== sec/4_experiments.tex =====

\subsection{Comparisons with previous results}
\label{sec:sota}

\paragraph{HumanML3D and KIT-ML.}
\Cref{tab:main} compares \ours with raw-motion, discrete-token, and continuous-latent generators.
\ours achieves the lowest FID on both benchmarks: 0.0253 on HumanML3D (5 RK4 steps, guidance 4.5) and 0.1214 on KIT-ML (12 RK4 steps, guidance 9.5).

\paragraph{Text consistency and distribution fidelity.}
R-Precision and MMDist assess text--motion alignment, while FID and Diversity characterize the generated motion distribution.
With standard CLIP sentence conditioning, \ours achieves the lowest FID on both benchmarks (\cref{tab:main}), highlighting the effectiveness of its motion representation under a standard generation setup.
Richer text features and more expressive text--motion interaction mechanisms offer complementary directions for finer-grained semantic control.

\paragraph{Qualitative comparison.}
\Cref{fig:qual} shows a challenging jump--spin--crouch sequence outside HumanML3D.
While prior methods exhibit distorted body configurations, \ours maintains coherent whole-body transitions.
Its dominant latent directions encode cross-joint coordination, which the flow recovers most accurately, while the decoder supplies joint-level detail (\cref{sec:recovery}).
A matched-generator comparison against a VAE latent appears in \cref{fig:supp_qual_vae}.

\begin{table}[t]
\centering\small
\renewcommand{\arraystretch}{1.5}
\setlength{\tabcolsep}{4pt}
\begin{tabular}{lccc}
\toprule
 & \multicolumn{2}{c}{Animators (1--5) $\uparrow$} & Users \\
\cmidrule(lr){2-3}\cmidrule(lr){4-4}
Method & Natural. & Text align. & prefer \ours \\
\midrule
MLD~\cite{chen2023mld} & 2.37 & 2.34 & 78\% \\
Light-T2M~\cite{zeng2025lightt2m} & 2.24 & 2.84 & 72\% \\
MoMask~\cite{guo2024momask} & 2.87 & 3.05 & 62\% \\
MotionHiFlow~\cite{li2026motionhiflow} & \underline{3.21} & \underline{3.50} & 64\% \\
\defaultrow \ours & \textbf{3.34} & \textbf{3.71} & -- \\
\bottomrule
\end{tabular}
\caption{\textbf{Human evaluation.} Two professional animators rate naturalness and text alignment on 19 prompts; 30 users provide pairwise preferences on 100 test prompts (\cref{sec:supp_animator}).}
\label{tab:animator}
\end{table}

\paragraph{Human evaluation.}
In blinded ratings of 19 prompts, two professional animators score \ours highest in naturalness and text alignment (\cref{tab:animator}), with paired 95\% confidence intervals excluding zero against MLD and Light-T2M on both criteria (\cref{tab:supp_animator}).
Across 100 randomly sampled test prompts, 30 users prefer \ours in 62--78\% of comparisons against MLD, Light-T2M, MoMask, and MotionHiFlow (\cref{tab:supp_pref}).

\paragraph{Model size.}
\label{sec:efficiency}
Inference uses a 58M DiT-M and a 12.8M decoder (70.7M total), with 5 RK4 steps requiring 20 network evaluations per sample.
On one RTX 5090, Motion-JEPA pre-training and DiT training take approximately 4 and 6 hours, respectively (\cref{tab:supp_efficiency,tab:supp_params}).

% ===== end sec/4_experiments.tex =====
% ===== sec/6_conclusion.tex =====
\section{Conclusion and Limitations}
\label{sec:conclusion}
We present \ours, a simple and effective framework that brings representation autoencoders to human motion generation.
By compressing frozen Motion-JEPA features into a compact latent space, \ours achieves state-of-the-art FID on HumanML3D and KIT-ML using standard CLIP sentence-level conditioning and a standard flow-matching Transformer.
Controlled experiments show that predictive pre-training yields representations that improve generation quality under the same generator configuration, and that the organization of motion structure across latent directions matters beyond dimensionality and condition number alone.

Our study focuses on the design of motion representation spaces, demonstrating that improving the representation modeled by the generator can substantially enhance the distributional fidelity of generated motions.
This direction is complementary to text-conditioning design: building on these representations, richer text features and more expressive text--motion interaction mechanisms offer promising avenues for finer-grained semantic control.
We hope \ours provides a simple, effective, and extensible framework for studying the relationship between motion representation and conditional generation.
% ===== end sec/6_conclusion.tex =====
% ===== sec/7_llm.tex =====
\section*{LLM Usage Statement}
\mbox{GPT-6 sol} (OpenAI) assisted with editing and refining author-produced text, code, and figures.
\mbox{GPT-6 sol} (OpenAI) and Claude (Anthropic) also provided reviewer-style feedback on the manuscript.
The authors take full responsibility for all submitted content.
% ===== end sec/7_llm.tex =====
{
    \small
    \bibliographystyle{ieeenat_fullname}
    \bibliography{refs}
}
\clearpage
% Included by main.tex; page numbering continues from the main paper.
\twocolumn[
  \centering
  \Large
  \textbf{\thetitle}\\
  \vspace{0.5em}Appendix\\
  \vspace{1.0em}
]
\setcounter{dbltopnumber}{4}
\setcounter{topnumber}{4}
\setcounter{totalnumber}{6}
\renewcommand{\dbltopfraction}{0.95}
\renewcommand{\topfraction}{0.95}
\renewcommand{\textfraction}{0.05}
\renewcommand{\floatpagefraction}{0.6}
\renewcommand{\dblfloatpagefraction}{0.6}
\setcounter{section}{0}\renewcommand{\thesection}{\Alph{section}}
\setcounter{table}{0}\renewcommand{\thetable}{S\arabic{table}}
\setcounter{figure}{0}\renewcommand{\thefigure}{S\arabic{figure}}
\setcounter{equation}{0}\renewcommand{\theequation}{S\arabic{equation}}
\renewcommand{\theHsection}{supp.\Alph{section}}
\renewcommand{\theHtable}{supp.\arabic{table}}
\renewcommand{\theHfigure}{supp.\arabic{figure}}
\renewcommand{\theHequation}{supp.\arabic{equation}}

This supplement provides additional analyses of MoRAE's latent space: representation controls (\cref{sec:supp_results}), cross-joint readout (\cref{sec:supp_analysis}), spectral recovery and coordinate transformations (\cref{sec:supp_geometry}), implementation and cost (\cref{sec:supp_impl}), human evaluation (\cref{sec:supp_animator}), and motion examples (\cref{sec:supp_qual}).

\section{Representation and compression controls}
\label{sec:supp_results}
\Cref{tab:supp_all} collects every representation, masking, and compression control under the same DiT-M training recipe, at both five and twelve RK4 steps, complementing the main-text ablations and the spectral overview in \cref{fig:spectral}a.
All numbers are means and 95\% CIs over 20 test-split repetitions with seeds 1234--1253; the five-step column corresponds to the comparisons in the main text.
% ===== sec/supp_tables.tex =====
\begin{table*}[!htbp]
\centering
\small
\setlength{\tabcolsep}{4.2pt}
\begin{tabular}{lrrcrc}
\toprule
 & & \multicolumn{2}{c}{5 RK4 steps} & \multicolumn{2}{c}{12 RK4 steps} \\
\cmidrule(lr){3-4}\cmidrule(lr){5-6}
Latent / generator & $d$ & $w$ & FID $\downarrow$ & $w$ & FID $\downarrow$ \\
\midrule
\multicolumn{6}{l}{\textit{Uncompressed spaces}} \\
raw features & 263 & 6.5 & 0.2502{\scriptsize$\pm$0.0062} & 6.5 & \textbf{0.0986}{\scriptsize$\pm$0.0042} \\
encoder features ($v$) & 768 & 4.5 & 0.3970{\scriptsize$\pm$0.0111} & 6.5 & \textbf{0.2192}{\scriptsize$\pm$0.0092} \\
encoder features ($x$) & 768 & 5.5 & 0.4372{\scriptsize$\pm$0.0133} & 6.5 & \textbf{0.2969}{\scriptsize$\pm$0.0102} \\
\midrule
\multicolumn{6}{l}{\textit{Reconstruction latents (same encoder architecture, no self-supervision)}} \\
AE ($\beta{=}0$) & 32 & 4.5 & \textbf{0.0349}{\scriptsize$\pm$0.0021} & 5.5 & 0.0453{\scriptsize$\pm$0.0024} \\
VAE ($\beta{=}10^{-4}$) & 32 & 4.5 & \textbf{0.0383}{\scriptsize$\pm$0.0019} & 4.5 & 0.0468{\scriptsize$\pm$0.0015} \\
VAE ($\beta{=}10^{-3}$) & 32 & 5.5 & 0.0375{\scriptsize$\pm$0.0013} & 6.5 & \textbf{0.0338}{\scriptsize$\pm$0.0013} \\
VAE ($\beta{=}10^{-2}$) & 32 & 5.5 & 0.0437{\scriptsize$\pm$0.0017} & 5.5 & \textbf{0.0383}{\scriptsize$\pm$0.0020} \\
\midrule
\multicolumn{6}{l}{\textit{Self-supervised objectives (32-D latent)}} \\
temporal mask only & 32 & 7.5 & 0.0660{\scriptsize$\pm$0.0043} & 6.5 & \textbf{0.0584}{\scriptsize$\pm$0.0025} \\
scrambled groups & 32 & 4.5 & \textbf{0.0377}{\scriptsize$\pm$0.0019} & 5.5 & 0.0477{\scriptsize$\pm$0.0029} \\
temporal + joint & 32 & 4.5 & \textbf{0.0271}{\scriptsize$\pm$0.0016} & 5.5 & 0.0390{\scriptsize$\pm$0.0021} \\
temporal 60\% + joint & 32 & 4.5 & \textbf{0.0393}{\scriptsize$\pm$0.0022} & 5.5 & 0.0435{\scriptsize$\pm$0.0022} \\
temporal + joint 50\% & 32 & 4.5 & \textbf{0.0374}{\scriptsize$\pm$0.0020} & 6.5 & 0.0604{\scriptsize$\pm$0.0035} \\
direct 32-D bottleneck (no compressor) & 32 & 5.5 & \textbf{0.0385}{\scriptsize$\pm$0.0018} & 5.5 & 0.0455{\scriptsize$\pm$0.0028} \\
\midrule
\multicolumn{6}{l}{\textit{\ours}} \\
\ours-16 & 16 & 4.5 & 0.0583{\scriptsize$\pm$0.0034} & 4.5 & \textbf{0.0501}{\scriptsize$\pm$0.0021} \\
\ours-32 & 32 & 4.5 & \textbf{0.0253}{\scriptsize$\pm$0.0015} & 5.5 & 0.0395{\scriptsize$\pm$0.0021} \\
\ours-32 whitened & 32 & 4.5 & \textbf{0.0411}{\scriptsize$\pm$0.0023} & 4.5 & 0.0449{\scriptsize$\pm$0.0025} \\
\ours-64 & 64 & 4.5 & \textbf{0.0262}{\scriptsize$\pm$0.0011} & 6.5 & 0.0438{\scriptsize$\pm$0.0026} \\
\midrule
\multicolumn{6}{l}{\textit{KIT-ML}} \\
\ours-32 (KIT-ML) & 32 & 7.5 & 0.1467{\scriptsize$\pm$0.0060} & 9.5 & \textbf{0.1214}{\scriptsize$\pm$0.0037} \\
VAE $\beta{=}10^{-4}$ (KIT-ML) & 32 & 4.5 & 0.1552{\scriptsize$\pm$0.0067} & 5.5 & \textbf{0.1443}{\scriptsize$\pm$0.0066} \\
\bottomrule
\end{tabular}
\caption{\textbf{Generation quality across representation controls} (HumanML3D unless noted; mean and 95\% CI over 20 repetitions). ($v$) and ($x$) denote velocity and clean-feature prediction.}
\label{tab:supp_all}
\end{table*}
% ===== end sec/supp_tables.tex =====
% ===== sec/supp_mae.tex =====
\begin{table*}[!htbp]
\centering\small
\setlength{\tabcolsep}{4pt}
\begin{tabular}{lccccccc}
\toprule
 & & & \multicolumn{3}{c}{R Precision$\uparrow$} & & \\
\cmidrule(lr){4-6}
Configuration & RK4 & FID $\downarrow$ & Top 1 & Top 2 & Top 3 & MMDist $\downarrow$ & Diversity $\to$ \\
\midrule
\ours-16 & 12 & 0.0501{\scriptsize$\pm$0.0021} & 0.535{\scriptsize$\pm$0.003} & 0.723{\scriptsize$\pm$0.003} & 0.815{\scriptsize$\pm$0.002} & 2.859{\scriptsize$\pm$0.008} & 9.669{\scriptsize$\pm$0.072} \\
\ours-64 & 5 & 0.0262{\scriptsize$\pm$0.0011} & 0.517{\scriptsize$\pm$0.002} & 0.706{\scriptsize$\pm$0.002} & 0.800{\scriptsize$\pm$0.002} & 2.976{\scriptsize$\pm$0.008} & 9.603{\scriptsize$\pm$0.070} \\
MAE & 12 & 0.0402\pmc{.0031} & 0.5215\pmc{.0033} & 0.7235\pmc{.0028} & 0.8148\pmc{.0022} & 2.8673\pmc{.0071} & 9.6687\pmc{.0727} \\
\bottomrule
\end{tabular}
\caption{\textbf{Additional latent widths and the MAE control} on HumanML3D (mean and 95\% CI over 20 evaluations).}
\label{tab:supp_mae}
\label{tab:supp_metrics}
\end{table*}
% ===== end sec/supp_mae.tex =====
% ===== sec/supp_width.tex =====
\paragraph{Latent width and the reconstruction-target control.}
\label{sec:supp_width}
MoRAE-64 reaches a comparable FID of $0.0262\pm0.0011$ with five RK4 steps; \cref{tab:supp_mae} gives its full metrics alongside MoRAE-16 and the MAE control.
Its spectral organization follows the same pattern as MoRAE-32: $\rho(\lambda,L_C/L_P)=0.925$ (\cref{tab:supp_spectral}).
The full width and sampler results appear in \cref{tab:supp_all}, with parameter counts in \cref{tab:supp_params}.
% ===== end sec/supp_width.tex =====

\section{Cross-joint structure in the latent}
\label{sec:supp_analysis}
\Cref{tab:supp_readout,tab:supp_budget} test the accessibility of cross-joint structure with linear and nonlinear readouts.
MoRAE-32 achieves the lowest $C$ error among the matched-width AE/VAE controls at both MLP budgets, showing that predictive pre-training encourages the latent to prioritize cross-joint coordination over joint-specific residuals.

\paragraph{Shared data and targets.}
All latent diagnostic analyses use non-mirrored HumanML3D clips (seed 20261025): 512 training clips for fitting the cross-joint predictors, a disjoint 1,024 for fitting probes and PCA, 256 for probe validation, and 512 for testing.
Clips have at least 40 frames and a full-sequence description; we take a centered crop of at most 196 frames, rounded to a multiple of four.
All autoencoders share the same clips, crops, and frame indices; conditional recovery uses the first full-sequence caption.
With motion features standardized by the dataset statistics, ridge regression ($10^{-3}$ regularization) predicts each joint block from the other joints at the same frame.
Concatenated predictions define $C(x)$, and $P(x)=x-C(x)$.
Readout squared errors are normalized by the target's total probe-training variance and averaged within clips.

\paragraph{Matched probe capacity.}
Each clip supplies 32 interior frames.
Ridge regularization is selected on validation data from $\{0,10^{-4},10^{-3},10^{-2},10^{-1},1\}$.
The two-layer MLP has width 128 and GELU activation, trained with AdamW (learning rate $10^{-3}$, weight decay $10^{-4}$, batch 2,048).
Training-normalized inputs are padded to width 64, matching probe parameter counts.
Three initializations are averaged; the 80- and 320-epoch caps both use validation selection and patience 10.
% ===== sec/supp_readout.tex =====
\begin{table*}[!htbp]
\centering\small
\setlength{\tabcolsep}{5pt}
\begin{tabular}{lcccc}
\toprule
Latent & $C$: linear & $C$: MLP 80 & $P$: linear & $P$: MLP 80 \\
\midrule
MoRAE-32 & 0.2192 [0.2009, 0.2400] & 0.1246 [0.1131, 0.1364] & 0.8432 [0.7717, 0.9153] & 0.7106 [0.6459, 0.7744] \\
AE-32 & 0.2310 [0.2121, 0.2504] & 0.1285 [0.1166, 0.1415] & 0.7566 [0.6889, 0.8273] & 0.5894 [0.5316, 0.6492] \\
VAE-32 & 0.2348 [0.2158, 0.2549] & 0.1288 [0.1166, 0.1418] & 0.7563 [0.6886, 0.8273] & 0.5893 [0.5318, 0.6486] \\
Whitened-32 & 0.2192 [0.2009, 0.2400] & 0.1245 [0.1130, 0.1363] & 0.8425 [0.7709, 0.9144] & 0.7113 [0.6472, 0.7741] \\
MoRAE-16 & 0.3350 [0.3113, 0.3594] & 0.1894 [0.1731, 0.2066] & 0.8972 [0.8235, 0.9720] & 0.8128 [0.7425, 0.8809] \\
MoRAE-64 & 0.1305 [0.1183, 0.1443] & 0.0822 [0.0751, 0.0898] & 0.7625 [0.6968, 0.8299] & 0.6537 [0.5949, 0.7130] \\
\bottomrule
\end{tabular}
\caption{\textbf{Readout results across latent spaces} (normalized MSE with 95\% clip-bootstrap CIs).}
\label{tab:supp_readout}
\end{table*}
% ===== end sec/supp_readout.tex =====
% ===== sec/supp_budget.tex =====
\begin{table*}[!htbp]
\centering\footnotesize
\setlength{\tabcolsep}{5pt}
\begin{tabular}{lcc}
\toprule
Latent & $C$: MLP 320 & $P$: MLP 320 \\
\midrule
AE-32 & 0.1111 [0.1002, 0.1227] & 0.5563 [0.5005, 0.6138] \\
VAE-32 & 0.1112 [0.1001, 0.1228] & 0.5562 [0.5012, 0.6135] \\
MoRAE-32 & 0.1078 [0.0975, 0.1184] & 0.6895 [0.6263, 0.7526] \\
\bottomrule
\end{tabular}
\caption{\textbf{Extended readout training budget} (320-epoch MLP, normalized MSE with 95\% clip-bootstrap CIs).}
\label{tab:supp_budget}
\end{table*}
% ===== end sec/supp_budget.tex =====
% ===== sec/supp_semantic_probe.tex =====
\paragraph{Action-recognition linear probe.}
\label{sec:supp_semantic}
The probe uses the official HumanML3D splits without mirrored clips and the preprocessing of the diagnostic study (Mean/Std normalization, centered crops of at most 196 frames), giving 11,207 training, 700 validation, and 2,104 test clips.
Action labels come from the full-motion captions: a clip receives one of 12 classes (walk, run, jump, sit, wave, throw, kick, dance, crouch, turn, punch, climb) when its captions match exactly one class keyword, which labels 943 test clips.
For each frozen encoder, the frame features used by the compressor (the normalized sum of the last four layers) are averaged over valid frames and standardized with training statistics; raw motion features are averaged in the same way.
A class-balanced multinomial logistic regression is fit on the training clips, with its regularization $C\in\{10^{-3},10^{-2},10^{-1},1\}$ selected by validation balanced accuracy.
\Cref{tab:supp_semantic} reports test balanced accuracy and paired differences over 1,000 clip bootstrap resamples.

\begin{table}[!htbp]
\centering\small
\setlength{\tabcolsep}{3.5pt}
\begin{tabular}{lccc}
\toprule
Frozen representation & $d$ & Bal.\ acc.\ (\%) & Motion-JEPA gain \\
\midrule
Motion-JEPA encoder & 768 & \textbf{82.1} & -- \\
AE encoder & 768 & 77.8 & $+4.2$ [1.0, 7.5] \\
VAE encoder & 768 & 78.8 & $+3.3$ [$-$0.2, 6.9] \\
Raw motion features & 263 & 70.9 & $+11.2$ [8.3, 14.5] \\
\bottomrule
\end{tabular}
\caption{\textbf{Linear action recognition from frozen features} on HumanML3D (12 classes; paired differences with 95\% bootstrap CIs).}
\label{tab:supp_semantic}
\end{table}
% ===== end sec/supp_semantic_probe.tex =====

\paragraph{Paired uncertainty.}
Intervals use 1,000 bootstrap resamples of test clips, averaging probe initializations and applicable flow/noise seeds within each clip before resampling.
For comparisons, we subtract per-clip errors first.
At the 320-epoch cap, the paired $C$-error difference is $-0.00333$ ($[-0.00516,-0.00169]$) between \ours-32 and AE-32, and $-0.00340$ ($[-0.00513,-0.00182]$) between \ours-32 and VAE-32.
These intervals describe the diagnostic clips for the fixed models; individual generator runs appear in \cref{tab:supp_all}.

\subsection{Interpreting the residual through context}
\label{sec:supp_context}
The residual $P(x)$ is defined relative to the same-frame linear predictor.
\Cref{tab:supp_context} tests how much of $P(x)$ can be predicted from a broader temporal context or nonlinear model when the target joint is excluded from every input frame.
Linear probes cover all 22 joints; nonlinear probes use root, spine, left/right feet, and left/right wrists (indices 0, 9, 10, 11, 20, 21).
We use 16 interior frames per clip, within-clip five-frame windows, matched input capacity, validation selection, and three MLP initializations; joint errors are weighted by their training target variances.
The nonlinear five-frame readout yields an NMSE of 0.7763 versus 0.7983 for a single frame (paired difference $-0.02200$, 95\% CI $[-0.04193,-0.00894]$).
Thus, even with nonlinear predictors and a five-frame window, nearly 78\% of the residual variance remains unpredictable from other joints, confirming that $P(x)$ primarily captures joint-specific variation rather than nonlinear or short-range cross-joint coordination.
% ===== sec/supp_context.tex =====
\begin{table*}[!htbp]
\centering\small
\setlength{\tabcolsep}{5pt}
\begin{tabular}{lccc}
\toprule
Predictor input & NMSE & 95\% interval & Mean baseline \\
\midrule
\multicolumn{4}{l}{\textit{Linear, 22 joints}} \\
Other joints, same frame & 0.8840 & [0.8095, 0.9649] & 0.9349 \\
Other joints, five frames & 0.8395 & [0.7679, 0.9156] & 0.9349 \\
\midrule
\multicolumn{4}{l}{\textit{MLP, 6 joints}} \\
Other joints, same frame & 0.7983 & [0.7221, 0.8869] & 0.9884 \\
Other joints, five frames & 0.7763 & [0.7049, 0.8560] & 0.9884 \\
\bottomrule
\end{tabular}
\caption{\textbf{Residual readout from other joints} (normalized MSE with 95\% clip-bootstrap CIs).}
\label{tab:supp_context}
\end{table*}
% ===== end sec/supp_context.tex =====

\section{Spectral organization and recovery}
\label{sec:supp_geometry}
\paragraph{Directional structure.}
For centered training PCA projection $q_i$ with variance $\lambda_i$, the loading of target $Y\in\{C,P\}$ is
\begin{equation}
 L_Y(i)=\frac{\|\operatorname{Cov}(q_i,Y)\|_2^2}
 {\lambda_i\operatorname{tr}\operatorname{Cov}(Y)}.
\end{equation}
This is the fraction of target variance explained by its single-direction linear readout.
\Cref{tab:supp_spectral} quantifies how this loading follows the spectrum: MoRAE places cross-joint structure in dominant directions and the joint-specific residual toward the trailing directions.
% ===== sec/supp_spectral.tex =====
\begin{table*}[!htbp]
\centering\small
\setlength{\tabcolsep}{5pt}
\begin{tabular}{lrrrr}
\toprule
Latent & $\lambda_1/\lambda_d$ & $\rho(\lambda,L_C)$ & $\rho(\lambda,L_P)$ & $\rho(\lambda,L_C/L_P)$ \\
\midrule
MoRAE-32 & 59.70 & 0.971 & -0.694 & 0.918 \\
AE-32 & 40.61 & 0.941 & -0.387 & 0.856 \\
VAE-32 & 27.02 & 0.933 & -0.293 & 0.815 \\
Whitened-32 & 4.65 & 0.304 & 0.033 & 0.229 \\
MoRAE-16 & 9.78 & 0.950 & -0.547 & 0.868 \\
MoRAE-64 & 252.19 & 0.971 & -0.740 & 0.925 \\
\bottomrule
\end{tabular}
\caption{\textbf{Training-spectrum diagnostics} in the normalized flow coordinates.}
\label{tab:supp_spectral}
\end{table*}
% ===== end sec/supp_spectral.tex =====

\paragraph{A common spectral protocol.}
\Cref{tab:supp_kappa_all} covers the 18 fixed input spaces.
For every space, we use the same 1,024 training clips and 32 interior frames per clip (32,768 vectors), with the exact per-channel normalization used to train its flow.
We center the vectors, compute covariance in float64, and report $\kappa=\lambda_1/\lambda_d$, without an eigenvalue floor or refitting channel normalization.
Raw motion and encoder features have numerical ranks 258/263 and 767/768 at tolerance $d\epsilon_{64}\lambda_1$.
Eigenvalue ratios near $10^{15}$ in \cref{fig:spectral}a mark numerical rank deficiency. Restricted to the numerically nonzero directions, the ratio is still $7.8\times10^{5}$ for raw motion and $4.5\times10^{4}$ for encoder features, compared with at most 252 for every compact latent. Below $10^{-3}\lambda_1$ there are 89 directions in raw motion and 312 in the encoder features, and none in any compact latent.
Five-step FIDs are reported in \cref{tab:supp_all}; the two-stage feature-decoder variant (which first decodes the 32-D latent to 768-D encoder features before decoding to motion) reaches FID 0.0309 at guidance 4.5 over 20 repetitions.
% ===== sec/supp_kappa_all.tex =====
\begin{table}[!htbp]
\centering\small
\setlength{\tabcolsep}{3pt}
\begin{tabular}{lrr}
\toprule
Input space & $d$ & $\kappa=\lambda_1/\lambda_d$ \\
\midrule
\multicolumn{3}{l}{\textit{Reconstruction objectives}} \\
AE ($\beta=0$) & 32 & 40.61 \\
VAE ($\beta=10^{-4}$) & 32 & 27.02 \\
VAE ($\beta=10^{-3}$) & 32 & 7.49 \\
VAE ($\beta=10^{-2}$) & 32 & 2.22 \\
\midrule
\multicolumn{3}{l}{\textit{Prediction task and compression}} \\
Temporal only & 32 & 21.77 \\
Scrambled groups & 32 & 46.97 \\
Temporal + joint & 32 & 52.22 \\
60\% temporal + joint & 32 & 39.26 \\
Temporal + 50\% joint & 32 & 67.81 \\
Direct 32-D bottleneck & 32 & 49.68 \\
Two-stage feature decoder & 32 & 41.95 \\
\midrule
\multicolumn{3}{l}{\textit{MoRAE width and coordinate controls}} \\
MoRAE-16 & 16 & 9.78 \\
MoRAE-32 & 32 & 59.70 \\
MoRAE-64 & 64 & 252.19 \\
Whitened MoRAE-32 & 32 & 4.65 \\
\midrule
\multicolumn{3}{l}{\textit{Uncompressed inputs}} \\
Raw motion features & 263 & \textit{rank deficient} \\
Encoder features ($v$) & 768 & \textit{rank deficient} \\
Encoder features ($x$) & 768 & \textit{rank deficient} \\
\bottomrule
\end{tabular}
\caption{\textbf{Condition numbers in the normalized flow coordinates.}}
\label{tab:supp_kappa_all}
\end{table}
% ===== end sec/supp_kappa_all.tex =====

\begin{figure*}[!htbp]
\centering
\includegraphics[width=\textwidth]{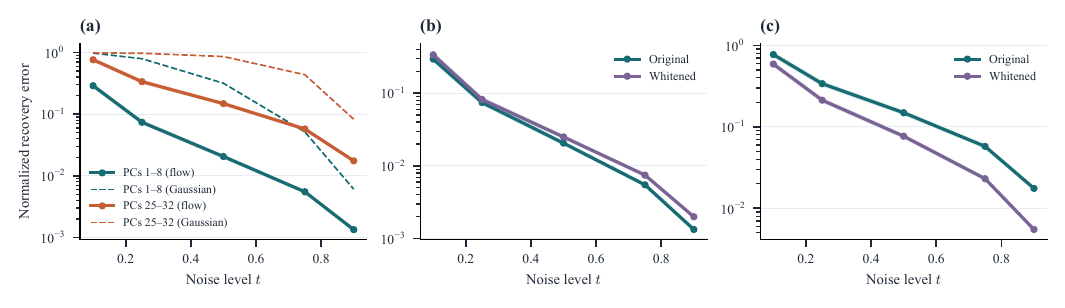}
\caption{\textbf{Directional recovery quartiles across noise levels.} (a) \ours-32 leading (PCs 1--8) and trailing (PCs 25--32) quartiles over three DiT seeds; dashed curves show the Gaussian reference. (b, c) Original and whitened models evaluated in the original PCA basis on the high-variance leading quartile (b, PCs 1--8) and low-variance trailing quartile (c, PCs 25--32) at guidance $w{=}1$ (95\% clip-bootstrap intervals).}
\label{fig:supp_directional}
\end{figure*}

\begin{figure*}[!htbp]
\centering
\includegraphics[width=\textwidth]{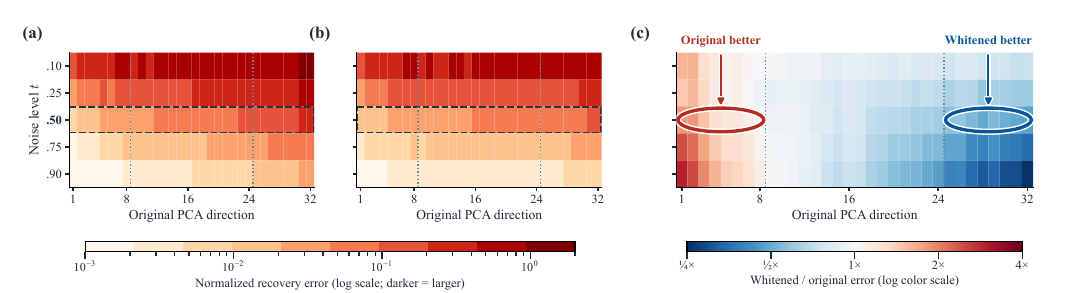}
\caption{\textbf{Per-direction recovery heatmaps under whitening.} Normalized recovery error across all 32 original PCA directions and noise levels $t\in\{0.10,0.25,0.50,0.75,0.90\}$ for (a) the original \ours-32 flow and (b) the whitened flow, together with (c) their error ratio: whitening improves the low-variance trailing directions (blue) at the expense of the coordination-dominant leading directions (red).}
\label{fig:supp_heatmap}
\end{figure*}

\paragraph{Directional recovery protocol.}
We use compressor posterior means standardized with each flow's saved training statistics.
At $t\in\{0.1,0.25,0.5,0.75,0.9\}$, \cref{eq:recovery} recovers known latents using the same three Gaussian noise seeds (11, 22, 33).
Errors in each PCA direction are divided by its training eigenvalue; the leading and trailing quartiles each contain eight directions at width 32.
\Cref{fig:supp_directional} shows the leading and trailing quartiles in panel (a); panels (b, c) and \cref{fig:supp_heatmap} compare the original and whitened models across all 32 original PCA coordinates.
The Gaussian linear-MMSE reference uses the training mean and covariance independently per frame, with the same corruptions.
Latent extraction and decoding use FP32. Mixed-precision flow evaluation was verified against FP32 across all ten diagnostic DiTs, five noise levels $t$, and two guidance settings on 32 clips (maximum mean-error change 0.111\%); the full 512-clip MoRAE-32 check gives 0.034\%.

% ===== sec/supp_spectral_theory.tex =====
\subsection{Spectral geometry and recovery demands}
\label{sec:supp_spectral_theory}
We derive the covariance-matched Gaussian reference used for directional recovery and connect it to the flow path.
This reference isolates the contribution of the latent spectrum: it uses centered frame-level latents and omits text and temporal context. The conditional DiT measurements in \cref{fig:supp_directional,fig:supp_heatmap} evaluate the trained model separately.

\paragraph{Directional recovery.}
In the PCA basis, let $z_i\sim\mathcal N(0,\lambda_i)$ and $\epsilon_i\sim\mathcal N(0,1)$ be independent. The linear path gives
\begin{equation}
 y_i=t z_i+(1-t)\epsilon_i,\qquad
 q_i(t)=\operatorname{Var}(y_i)=t^2\lambda_i+(1-t)^2.
\end{equation}
Gaussian conditioning yields
\begin{align}
 \mathbb E[z_i\mid y_i]&=\frac{t\lambda_i}{q_i(t)}y_i,\\
 \operatorname{Var}(z_i\mid y_i)&=\frac{\lambda_i(1-t)^2}{q_i(t)}.
\end{align}
Dividing the conditional variance by $\lambda_i$ gives the normalized error in \cref{eq:spectral_recovery}.
It decreases with the directional signal-to-noise ratio $t^2\lambda_i/(1-t)^2$.
For a positive-definite covariance and $0<t<1$,
\begin{equation}
 \frac{\mathrm{SNR}_1(t)}{\mathrm{SNR}_d(t)}=\kappa.
\end{equation}
Solving $\mathrm{SNR}_i(t_i^*)=1$ gives $t_i^*=1/(1+\sqrt{\lambda_i})$ and
\begin{equation}
 \frac{t_d^*/(1-t_d^*)}{t_1^*/(1-t_1^*)}=\sqrt{\kappa}.
\end{equation}
Thus, spectral spread separates the times at which different directions become informative. When $\lambda_i$ is small, that direction's transition sits close to the data endpoint, at $1-t_i^*=\sqrt{\lambda_i}/(1+\sqrt{\lambda_i})$.

\paragraph{Target-weighted recovery at a fixed spectrum.}
When the eigenvalue multiset is held fixed and original PCA direction $i$ is assigned variance $\nu_i$, the normalized Gaussian reference error after inverting back to the original coordinates is $e_{i,\nu}^*(t)=\big(1+t^2\nu_i/(1-t)^2\big)^{-1}$.
Writing the linear readout of $C$ from the centered PCA projections $q_i$ as $\widetilde C=\mu_C+\sum_i b_i q_i$ with $b_i=\operatorname{Cov}(q_i,C)/\lambda_i$, independence across Gaussian PCA coordinates yields $\mathcal{E}_{C,\mathrm{lin}}^*(t;\nu)=\sum_{i=1}^d L_C(i)\,e_{i,\nu}^*(t)$ in \cref{eq:rearrangement}.
Holding the eigenvalue multiset fixed preserves $\kappa$ and the aggregate unweighted Gaussian recovery error $\sum_{i=1}^d e_{i,\nu}^*(t)$, while reallocating recovery accuracy across the original PCA directions.
Because $e_{i,\nu}^*(t)$ decreases strictly with $\nu_i$ for $0<t<1$, the rearrangement inequality implies that pairing larger $L_C(i)$ with larger $\nu_i$ ($C$-aligned) minimizes $\mathcal{E}_{C,\mathrm{lin}}^*(t;\nu)$, whereas the reverse pairing maximizes it: the spectrum sets the recovery scales, and the cross-joint loadings determine which motion structure those scales serve.

\paragraph{Finite-step flow dynamics.}
For this Gaussian reference, the optimal flow velocity is
\begin{equation}
 v_i^*(y_i,t)=\mathbb E[z_i-\epsilon_i\mid y_i]
 =\frac{t\lambda_i-(1-t)}{q_i(t)}y_i.
 \label{eq:supp_gaussian_velocity}
\end{equation}
The coefficient changes from $-1$ at $t=0$ to $1$ at $t=1$ for $\lambda_i>0$.
For a small eigenvalue, the factor $q_i(t)^{-1}$ creates rapidly varying dynamics near the endpoint. A finite-capacity velocity model and a fixed-step solver must represent and resolve these direction-dependent scales.
This provides a mechanism linking the spectrum to generation difficulty under a finite computational budget. Because generation quality also depends on the full eigenvalue distribution, text conditioning, and the ODE solver, the condition number alone does not determine a universal FID threshold.

\paragraph{From coordinates to decoded motion.}
An invertible coordinate change $z'=Az$ preserves information when decoding uses $D_A(z')=D(A^{-1}z')$.
For a generated endpoint error $\delta'$ in the transformed coordinates, the corresponding error in the original coordinates is exactly $A^{-1}\delta'$.
Locally, with $D$ including the inverse flow normalization,
\begin{equation}
 D_A(Az+\delta')-D(z)\approx J_D(z)A^{-1}\delta'.
\end{equation}
For an evaluator feature map $\phi$, the first-order feature perturbation is $J_{\phi\circ D}(z)A^{-1}\delta'$.
Because FID is computed on evaluator features of decoded motion, both the direction of the generated error and the response of the decoder and evaluator to that direction shape the feature statistics.
Directional recovery and motion readout are therefore complementary to a single condition number: one-step latent error does not map one-to-one to free-generation FID.
% ===== end sec/supp_spectral_theory.tex =====

\subsection{Reconstruction-preserving whitening transformation}
\label{sec:supp_white}
Let $\mu$ and $\Sigma=U\Lambda U^\top$ be the training latent mean and covariance, and define $A=\Lambda^{-1/2}U^\top$, $z_w=A(z-\mu)$.
The compressor mean head $m=Wh+b$ becomes $W'=AW$, $b'=A(b-\mu)$; the decoder input $u=W_dz+b_d$ becomes $W'_d=W_dA^{-1}$, $b'_d=b_d+W_d\mu$.
Thus $D'(z_w)=D(z)$, with maximum absolute reconstruction difference $5.13\times10^{-5}$ on all 512 test clips.
Each flow retains its recorded normalization; the effective diagnostic condition numbers are 59.70 and 4.65.

Whitened predictions are inverted, standardized by the original model's statistics, and projected onto its PCA basis.
This gives the common-coordinate comparison in \cref{fig:supp_directional,fig:supp_heatmap}.
\Cref{tab:supp_recovery} reports the decoded $C/P$ errors against real motion with a matched DiT training seed, integrated via the trapezoidal rule over the five noise levels $t\in[0.1,0.9]$ and normalized by the interval length $0.8$.
Free generation uses five RK4 steps and guidance 4.5 for both coordinate systems (\cref{tab:white}).
Reconstruction is identical while directional recovery shifts, so the FID gap is not lost information; the next subsection separates the coordinate change from the change in loss weights.
% ===== sec/supp_recovery.tex =====
\begin{table}[!htbp]
\centering\small
\setlength{\tabcolsep}{3pt}
\begin{tabular}{rcccc}
\toprule
& \multicolumn{2}{c}{$C$ error} & \multicolumn{2}{c}{$P$ error} \\
\cmidrule(lr){2-3}\cmidrule(lr){4-5}
Guidance $w$ & Original & Whitened & Original & Whitened \\
\midrule
1 & 0.08539 & 0.08546 & 0.46466 & 0.42556 \\
4.5 & 0.22800 & 0.25645 & 0.72310 & 0.67490 \\
\bottomrule
\end{tabular}
\caption{\textbf{Recovery allocation under whitening.} Integrated normalized error against real motion; matched DiT training seed, 512 clips, and three shared noise draws.}
\label{tab:supp_recovery}
\end{table}
% ===== end sec/supp_recovery.tex =====

\subsection{Coordinate and loss-metric controls}
\label{sec:supp_coord_loss}
We evaluate a $2\times 2$ design crossing generator coordinates with the training loss metric.
Let $q$ denote the original normalized flow latent and $q_w = Bq + b$ denote its whitened counterpart, including the per-channel normalization used by each generator.
Writing the velocity error in original coordinates as $e = \hat{v} - v$, the original- and whitened-space loss metrics are
\begin{equation}
\mathcal{L}_{o,o}=\mathbb{E}\|e\|_2^2/d,
\qquad
\mathcal{L}_{o,w}=\mathbb{E}\|Be\|_2^2/d.
\end{equation}
For the velocity error $e_w$ in whitened coordinates, the corresponding losses are
\begin{equation}
\mathcal{L}_{w,o}=\mathbb{E}\|B^{-1}e_w\|_2^2/d,
\qquad
\mathcal{L}_{w,w}=\mathbb{E}\|e_w\|_2^2/d,
\qquad d=32.
\end{equation}
Each row uses its own coordinate system and isotropic Gaussian noise process; predictions are mapped back through the corresponding inverse transform before the frozen motion decoder.
We reuse the two diagonal checkpoints and train the two off-diagonal configurations with the matched DiT training seed, a global batch size of 64, and the same 500-epoch schedule (resumed from saved optimizer and EMA states across distributed continuations on one, four, and two GPUs while preserving the global batch and schedule).
All evaluations use the epoch-500 EMA checkpoint, five RK4 steps, and guidance $w{=}4.5$ on the 4,382 finite-feature test clips.
\Cref{tab:supp_coord_loss} uses the 15 sampling seeds completed for every cell (1234--1242 and 1244--1249), which is why its original/original FID is 0.0244 rather than the full-split 0.0253.
Over all 20 seeds on the same 4,382 clips, the other three cells are $0.0650\pm0.0033$ ($\mathcal{L}_{o,w}$), $0.0410\pm0.0025$ ($\mathcal{L}_{w,o}$), and $0.0406\pm0.0019$ ($\mathcal{L}_{w,w}$); the last of these matches the whitened generator re-scored on this subset (\cref{tab:supp_kappa_ctrl}), against 0.0411 on the full split (\cref{tab:white}).

\begin{table}[!htbp]
\centering\small
\setlength{\tabcolsep}{2pt}
\begin{tabular}{lcc}
\toprule
Generator coordinates & Original-space loss & Whitened-space loss \\
\midrule
Original & \textbf{0.0244}\pmc{.0011} & 0.0659\pmc{.0037} \\
Whitened & 0.0407\pmc{.0026} & 0.0416\pmc{.0016} \\
\bottomrule
\end{tabular}
\caption{\textbf{Generator coordinates and loss metric} (FID $\downarrow$ on HumanML3D; mean and 95\% CI over 15 shared sampling seeds).}
\label{tab:supp_coord_loss}
\end{table}

The native original-coordinate, original-metric configuration achieves the best distribution fidelity among the four interfaces.
Restoring the original-space metric in whitened coordinates ($\mathcal{L}_{w,o}$) yields a similar FID to the whitened-space metric ($\mathcal{L}_{w,w}$) while preserving the gap to the original interface.
Loss reweighting therefore does not account for the full coordinate-transformation effect; the fixed-spectrum experiment in \cref{sec:supp_iso} complements this control by isolating the assignment of cross-joint information to variance directions.

\subsection{Isospectral alignment intervention}
\label{sec:supp_iso}
To test directional alignment without changing the eigenvalue spectrum, we keep the frozen \ours-32 compressor, decoder, and flow normalization statistics ($\mu$ and covariance $\Sigma=U\operatorname{diag}(\lambda)U^\top$ fit on the 1,024 diagnostic training clips, 32 interior frames per clip).
Let $q=U^\top(z-\mu)$ denote the original PCA projections and $L_C(i)$ their training-set cross-joint readout loadings.
We form two target variance vectors $\nu$ by assigning the exact same eigenvalue multiset $\{\lambda_i\}_{i=1}^{32}$ to the original PCA directions in descending order of $L_C(i)$ ($C$-aligned) or ascending order of $L_C(i)$ (Reverse-aligned), via the invertible linear transformation
\begin{equation}
 A_\nu = U\operatorname{diag}\!\Big(\sqrt{\nu_i/\lambda_i}\Big)U^\top,\qquad
 z_\nu = A_\nu(z-\mu).
 \label{eq:supp_iso_map}
\end{equation}
Thus $\operatorname{Cov}(z_\nu)=U\operatorname{diag}(\nu)U^\top$ preserves the full eigenvalue multiset, trace, and condition number ($\kappa=59.7041$), with a maximum relative spectral error of $7.31\times 10^{-15}$ ($C$-aligned) and $5.52\times 10^{-15}$ (Reverse-aligned).
At decoding, we apply $z=A_\nu^{-1}z_\nu+\mu$ before the unchanged inverse flow normalization and frozen motion decoder; the FP32 latent round-trip error stays below $10^{-6}$, and the GPU decoded-motion relative MSE is $1.89\times 10^{-13}$ ($C$-aligned) and $2.13\times 10^{-13}$ (Reverse-aligned).
Both spaces train a DiT-M with the identical recipe, initialization, batch order, and epoch-500 EMA checkpoint, using standard isotropic Gaussian noise and unweighted flow MSE in their respective coordinates.

\Cref{tab:supp_iso_gen} reports all six generation metrics on HumanML3D over 20 sampling repetitions (seeds 1234--1253) at five RK4 steps and guidance $w{=}4.5$.
\Cref{tab:supp_iso_rec} integrates the normalized decoded $C$ and $P$ recovery errors over $t\in\{0.1,0.25,0.5,0.75,0.9\}$ on the 512 test clips and three shared noise draws (trapezoidal rule divided by $0.8$, evaluated in FP32).
Each error is evaluated both against the shared autoencoder reconstruction $x_{\mathrm{rec}}$ (isolating generator recovery error) and against ground-truth motion $x$, with 1,000-sample paired clip-bootstrap 95\% confidence intervals for $C$-aligned minus Reverse-aligned.
Across both guidance settings and both references, $C$-aligned lowers both $C$ and $P$ recovery errors, and every paired 95\% confidence interval lies strictly below zero.

\begin{table*}[!htbp]
\centering\small
\setlength{\tabcolsep}{5pt}
\begin{tabular}{lccccccc}
\toprule
 & & & \multicolumn{3}{c}{R Precision$\uparrow$} & & \\
\cmidrule(lr){4-6}
Spectrum assignment & $\kappa$ & FID $\downarrow$ & Top 1 & Top 2 & Top 3 & MMDist $\downarrow$ & Diversity $\to$ \\
\midrule
$C$-aligned & 59.70 & \textbf{0.0279}\pmc{.0016} & 0.5148\pmc{.0025} & 0.7025\pmc{.0029} & 0.7977\pmc{.0029} & 2.9726\pmc{.0087} & 9.5366\pmc{.0963} \\
Reverse-aligned & 59.70 & 0.0589\pmc{.0030} & 0.5190\pmc{.0036} & 0.7094\pmc{.0031} & 0.8063\pmc{.0026} & 2.9155\pmc{.0107} & 9.5850\pmc{.0852} \\
\bottomrule
\end{tabular}
\caption{\textbf{Full generation metrics for the isospectral alignment intervention} on HumanML3D (five RK4 steps, $w{=}4.5$).}
\label{tab:supp_iso_gen}
\end{table*}

\begin{table*}[!htbp]
\centering\small
\setlength{\tabcolsep}{2.5pt}
\begin{tabular}{rllcccc}
\toprule
Guidance $w$ & Target reference & Component & $C$-aligned $\downarrow$ & Reverse-aligned $\downarrow$ & Rel.\ reduction & Paired diff.\ [95\% CI] \\
\midrule
1.0 & Autoencoder recon.\ $x_{\mathrm{rec}}$ & $C$ error & \textbf{0.07417} & 0.10299 & 28.0\% & $-0.02882$ [$-0.03258$, $-0.02517$] \\
1.0 & Autoencoder recon.\ $x_{\mathrm{rec}}$ & $P$ error & \textbf{0.29132} & 0.31508 & 7.5\% & $-0.02376$ [$-0.03026$, $-0.01658$] \\
4.5 & Autoencoder recon.\ $x_{\mathrm{rec}}$ & $C$ error & \textbf{0.21838} & 0.34707 & 37.1\% & $-0.12869$ [$-0.15286$, $-0.10790$] \\
4.5 & Autoencoder recon.\ $x_{\mathrm{rec}}$ & $P$ error & \textbf{0.56166} & 0.65146 & 13.8\% & $-0.08980$ [$-0.10647$, $-0.07336$] \\
\midrule
1.0 & Ground-truth motion $x$ & $C$ error & \textbf{0.08530} & 0.11387 & 25.1\% & -- \\
1.0 & Ground-truth motion $x$ & $P$ error & \textbf{0.46401} & 0.48535 & 4.4\% & -- \\
4.5 & Ground-truth motion $x$ & $C$ error & \textbf{0.22862} & 0.35722 & 36.0\% & -- \\
4.5 & Ground-truth motion $x$ & $P$ error & \textbf{0.72419} & 0.81143 & 10.8\% & -- \\
\bottomrule
\end{tabular}
\caption{\textbf{Decoded $C$ and $P$ recovery under the isospectral alignment intervention} (time-integrated normalized error and 95\% bootstrap CIs).}
\label{tab:supp_iso_rec}
\end{table*}

\subsection{Conditioning at fixed width}
\label{sec:supp_fixedwidth}
\begin{table}[t]
\centering\small
\setlength{\tabcolsep}{4pt}
\begin{tabular}{lcccc}
\toprule
32-D coordinates & $\kappa$ & Guid. & FID $\downarrow$ & R@1 $\uparrow$ \\
\midrule
Original$^\ddagger$ & 59.70 & 4.5 & \textbf{0.0252}\pmc{.0019} & 0.516\pmc{.009} \\
Whitened & 4.65 & 4.5 & 0.0406\pmc{.0019} & 0.520\pmc{.003} \\
\midrule
Stretched & $10^2$ & 3.5 & 0.0476\pmc{.0032} & 0.529\pmc{.003} \\
Stretched & 133 & 3.5 & 0.0408\pmc{.0032} & 0.532\pmc{.003} \\
Stretched & 178 & 3.5 & 0.0592\pmc{.0035} & 0.525\pmc{.004} \\
Stretched & 237 & 3.5 & 0.0790\pmc{.0043} & 0.526\pmc{.004} \\
Stretched & 316 & 3.5 & 0.0963\pmc{.0040} & 0.524\pmc{.003} \\
Stretched & 562 & 3.5 & 0.2010\pmc{.0067} & 0.519\pmc{.003} \\
Stretched & 750 & 3.5 & 0.2701\pmc{.0083} & 0.515\pmc{.003} \\
Stretched & $10^3$ & 3.5 & 0.3742\pmc{.0093} & 0.513\pmc{.003} \\
Stretched & 1{,}778 & 3.5 & 0.960\pmc{.017} & 0.492\pmc{.003} \\
Stretched & $10^4$ & 3.5 & 8.870\pmc{.051} & 0.365\pmc{.002} \\
Stretched & $10^5$ & 3.5 & 35.52\pmc{.08} & 0.175\pmc{.002} \\
Stretched & $10^6$ & 5.5 & 53.04\pmc{.06} & 0.050\pmc{.001} \\
\bottomrule
\end{tabular}
\caption{\textbf{Conditioning at fixed width} on HumanML3D (\ours-32 under invertible geometric-spectrum maps; $^\ddagger$3 repetitions).}
\label{tab:supp_kappa_ctrl}
\end{table}
The uncompressed spaces differ from the compact latents in both conditioning and width.
To separate the two, we keep \ours-32 and change only its conditioning.
Let $V\Lambda V^\top$ be the covariance of the standardized training latents.
The flow is trained on $z'=Az$ with $A=RSW$, where $W=V\Lambda^{-1/2}V^\top$ whitens, $S=\mathrm{diag}(s_i)$ with $s_i=\kappa^{-(i-1)/(2(d-1))}$ sets a geometric spectrum from 1 to $1/\kappa$, and $R$ is a fixed random rotation, so the small directions do not align with coordinate axes.
Generated latents are mapped back by $A^{-1}$ in float64 before the unchanged decoder, so reconstruction is identical for every $\kappa$.
Each flow uses the DiT-M recipe of \cref{tab:width} (500 epochs, matched training seed, TF32 matrix multiplication). Guidance is selected on the validation split by a sweep over $\{3.5, 4.5, 5.5, 6.5\}$; \cref{tab:supp_kappa_ctrl} reports five RK4 steps and 20 test-split repetitions at that guidance.
Two test clips (007975 and its mirror) contain non-finite features in our copy of HumanML3D, so this control uses the remaining 4,382 clips.
Re-scored on those clips, the original \ours-32 generator has FID 0.0252 over three repetitions (versus 0.0253 over 20 on the full split), and the whitened generator has FID 0.0406 over 20 (versus 0.0411 on the full split).

FID stays at the $\kappa=10^2$ level through $\kappa=133$ and rises beyond it; $\kappa=178$ is the first setting whose 95\% CI no longer overlaps with that of $\kappa=10^2$ (\cref{tab:supp_kappa_ctrl}).
FID matches raw motion's 0.25 (\cref{tab:supp_all}) between $\kappa=562$ and 750, near $\kappa=700$ by log-linear interpolation, and grows by more than two orders of magnitude by $10^4$.
R@1 stays within 0.02 of the $\kappa=10^2$ latent up to $10^3$ and falls to 0.050 at $10^6$, near the $1/32$ random-retrieval chance level (32-candidate pool).
Learned compact latents span $\kappa$ from 2.2 to 252 (\cref{tab:supp_kappa_all}), so the $\kappa$ at which this synthetic curve rises is not a universal cutoff for learned latents.
Moreover, the $\kappa=10^2$ latent generates worse than the original coordinates, just as the whitened latent does (\cref{tab:white,tab:supp_kappa_ctrl}): the learned variance hierarchy aids generation beyond the scalar condition number alone.

\section{Implementation and computational cost}
\label{sec:supp_impl}
\Cref{tab:supp_impl} specifies the three training stages.
AE/VAE controls share the encoder architecture, jointly trained with the bottleneck and decoder from random initialization; their KL weights distinguish the reconstruction variants.
Uncompressed controls use the same DiT-M trunk in standardized raw or 768-D feature space, with a four-block motion decoder for encoder features; the clean-feature-prediction variant ($x$) predicts the clean target $\hat{x}$ and converts it to velocity $(\hat{x}-x_t)/(1-t)$ during ODE integration.
Temporal masks hide contiguous spans of 15--40 frames. Scrambled groups preserve the anatomical group sizes with randomly assigned channels; the direct 32-D bottleneck variant attaches a 32-D linear projection directly to the encoder.
Encoder, compressor, and decoder training are separate from the matched DiT training budget.
% ===== sec/supp_impl_details.tex =====
\begin{table}[!htbp]
  \centering
  \small
  \setlength{\tabcolsep}{3pt}
  \begin{tabular}{p{0.22\columnwidth}p{0.72\columnwidth}}
    \toprule
    \multicolumn{2}{l}{\textit{Motion-JEPA (\cref{sec:jepa})}} \\
    encoder & 8 transformer blocks, width 768, 12 heads, MLP ratio 4, rotary positions, one token per frame \\
    predictor & 6 blocks, width 384, 6 heads; output projected to 768 \\
    masking & 30\% of the joints (7 of 22; 6 of 21 on KIT-ML), all channels, all frames; learned mask values \\
    target & layer-normalized output of the EMA teacher, momentum 0.996$\to$1; smooth-$\ell_1$ on every valid frame \\
    optimization & AdamW ($\beta$ 0.9/0.95), lr $2{\cdot}10^{-4}\to10^{-5}$ cosine, 20 warm-up epochs, weight decay 0.05, batch 256, 600 epochs, clips of 24--200 frames \\
    features & layer norm of the sum of the last four teacher layers, $H\in\mathbb{R}^{T\times768}$ \\
    \midrule
    \multicolumn{2}{l}{\textit{Compact motion latent (\cref{sec:code})}} \\
    compressor $E_\psi$ & one multi-head attention layer (24 heads, width 3072) with residual and layer norm, linear head to mean and log-variance of $z\in\mathbb{R}^{T\times d}$ \\
    decoder $D_\omega$ & linear input, 4 transformer blocks (width 512, 8 heads, rotary positions), layer norm, linear output to 263 (251) features; 12.8M parameters \\
    loss & MSE on all motion features $+\,10^{-4}$ KL \\
    optimization & AdamW ($\beta$ 0.9/0.95), lr $2{\cdot}10^{-4}\to10^{-5}$ cosine, 10 warm-up epochs, weight decay 0.01, batch 384, 400 epochs; checkpoint with the lowest validation reconstruction error \\
    \midrule
    \multicolumn{2}{l}{\textit{Latent flow matching (\cref{sec:flow})}} \\
    DiT-M & 12 blocks, width 512, 8 heads, feed-forward 2048, dropout 0.1, adaptive layer norm on timestep $+$ CLIP ViT-B/32 sentence embedding; 58.0M parameters \\
    input & compressor posterior means, standardized per channel with training-set statistics; one token per frame, up to 196 frames \\
    optimization & AdamW ($\beta$ 0.9/0.99), lr $2{\cdot}10^{-4}$ cosine to 1\%, 2000 warm-up iterations, weight decay $10^{-5}$, batch 64, 500 epochs, EMA 0.9999 (used for sampling), text dropped with probability 0.1 \\
    sampling & probability-flow ODE, RK4 on a uniform grid of 5 (HumanML3D) or 12 (KIT-ML) steps, guidance $w{=}4.5$ (HumanML3D) and $w{=}9.5$ (KIT-ML) \\
    \bottomrule
  \end{tabular}
  \caption{\textbf{Architecture and training} of \ours. Every stage is trained on the training split of its dataset only, on a single GPU.}
  \label{tab:supp_impl}
\end{table}

% ===== sec/supp_efficiency.tex =====
\begin{table}[!htbp]
\centering\small
\setlength{\tabcolsep}{3pt}
\begin{tabular}{lrr}
\toprule
Shared module & HumanML3D & KIT-ML \\
\midrule
Motion-JEPA encoder & 56.91 & 56.90 \\
JEPA predictor & 11.24 & 11.24 \\
EMA teacher & 56.91 & 56.90 \\
Frozen CLIP text encoder & 63.43 & 63.43 \\
\midrule
\multicolumn{3}{l}{\textit{Recorded runtime on one RTX 5090}} \\
Motion-JEPA training & \multicolumn{2}{r}{$\approx$4 h} \\
DiT training & \multicolumn{2}{r}{$\approx$6 h} \\
Inference & \multicolumn{2}{r}{$\approx$0.06 s} \\
\bottomrule
\end{tabular}
\caption{\textbf{Shared model size (M) and runtime} on one RTX 5090.}
\label{tab:supp_efficiency}
\end{table}
% ===== end sec/supp_efficiency.tex =====
\begin{table}[!htbp]
\centering\small
\setlength{\tabcolsep}{3pt}
\begin{tabular}{lrrr}
\toprule
Dataset / width & Compressor & Decoder & DiT \\
\midrule
HumanML3D / 16 & 9.47 & 12.75 & 57.91 \\
HumanML3D / 32 & 9.50 & 12.76 & 57.93 \\
HumanML3D / 64 & 9.55 & 12.78 & 57.96 \\
KIT-ML / 32 & 9.50 & 12.76 & 57.93 \\
\bottomrule
\end{tabular}
\caption{\textbf{Width-dependent parameter counts (M).}}
\label{tab:supp_params}
\end{table}
% ===== end sec/supp_impl_details.tex =====

\paragraph{Evaluation and sampling.}
Both datasets follow the standard evaluation protocol~\cite{guo2022humanml3d}.
Guidance, the number of RK4 steps, and the generator checkpoint are selected on the validation split.
For each generator and step count, guidance is chosen by a one-repetition sweep in increments of 1.0.
Compressor and decoder checkpoints use the lowest validation reconstruction error.
Reported metrics are test-split evaluations with 20 sampling repetitions.
The main settings are five RK4 steps and guidance 4.5 on HumanML3D, and twelve steps and guidance 9.5 on KIT-ML.
Each RK4 step uses four guided velocity evaluations, batching conditional and unconditional branches.
At generation, only the flow, motion decoder, and frozen text tower are used; \cref{tab:supp_efficiency,tab:supp_params} separate shared and width-dependent module sizes.

\section{Human evaluation}
\label{sec:supp_animator}
\paragraph{Professional animators.}
Two professional animators independently rated the motions of five methods on 19 prompts that we wrote (\cref{sec:supp_qual}).
For each prompt, one synchronized video shows the five generated motions side by side as body meshes under a shared camera, labeled only as Method A to Method E.
The animator scores every method on motion naturalness and on alignment with the text prompt (1 = very poor, 5 = very good), can step through the video frame by frame, and is asked not to discuss which method is which.
All methods are generated as for \cref{fig:qual}: the prior methods use their official code, released checkpoints, and documented sampling settings, \ours uses the 32-D model with five RK4 steps and guidance $w{=}4.5$, and no method receives post-processing.
Every method receives 38 ratings per criterion.
\Cref{tab:supp_animator} reports mean scores and the paired difference between \ours and each method; intervals are 95\% bootstrap intervals over 10,000 resamples of the 19 prompts after averaging the two animators.

\begin{table*}[!htbp]
\centering\small
\setlength{\tabcolsep}{5pt}
\begin{tabular}{lcccc}
\toprule
 & \multicolumn{2}{c}{Naturalness $\uparrow$} & \multicolumn{2}{c}{Text alignment $\uparrow$} \\
\cmidrule(lr){2-3}\cmidrule(lr){4-5}
Method & Mean & \ours $-$ method & Mean & \ours $-$ method \\
\midrule
MLD & 2.37 [1.89, 2.84] & $+0.97$ [$+0.34$, $+1.61$] & 2.34 [1.87, 2.82] & $+1.37$ [$+0.74$, $+1.95$] \\
Light-T2M & 2.24 [1.95, 2.55] & $+1.11$ [$+0.82$, $+1.42$] & 2.84 [2.39, 3.32] & $+0.87$ [$+0.32$, $+1.45$] \\
MoMask & 2.87 [2.39, 3.34] & $+0.47$ [$-0.03$, $+0.97$] & 3.05 [2.50, 3.58] & $+0.66$ [$+0.00$, $+1.29$] \\
MotionHiFlow & 3.21 [2.76, 3.61] & $+0.13$ [$-0.32$, $+0.55$] & 3.50 [2.95, 4.03] & $+0.21$ [$-0.32$, $+0.76$] \\
\ours & \textbf{3.34} [2.95, 3.74] & -- & \textbf{3.71} [3.21, 4.18] & -- \\
\bottomrule
\end{tabular}
\caption{\textbf{Animator ratings} (1--5 scale; means and paired differences with 95\% prompt-bootstrap intervals).}
\label{tab:supp_animator}
\end{table*}

\paragraph{User preference study.}
We draw 100 prompts at random from the HumanML3D test split.
Each of 30 users is assigned 20 of these prompts at random and, for each prompt, compares the motion of \ours with the motion of each prior method in pairs, choosing the one preferred overall.
Method identities are hidden, and the left/right placement of the two motions is randomized.
This gives 600 comparisons per prior method.
\Cref{tab:supp_pref} reports the share of comparisons in which \ours is preferred, with 95\% Wilson intervals.

\begin{table*}[!htbp]
\centering\small
\setlength{\tabcolsep}{5pt}
\begin{tabular}{lcc}
\toprule
\ours versus & Prefer \ours & 95\% interval \\
\midrule
MLD & 78\% & [74.5, 81.1] \\
Light-T2M & 72\% & [68.3, 75.4] \\
MotionHiFlow & 64\% & [60.1, 67.7] \\
MoMask & 62\% & [58.1, 65.8] \\
\bottomrule
\end{tabular}
\caption{\textbf{Pairwise user preference} (30 users, 600 comparisons per method).}
\label{tab:supp_pref}
\end{table*}

\section{Generated motion examples}
\label{sec:supp_qual}
\begin{figure*}[!htbp]
\centering
\includegraphics[width=\textwidth]{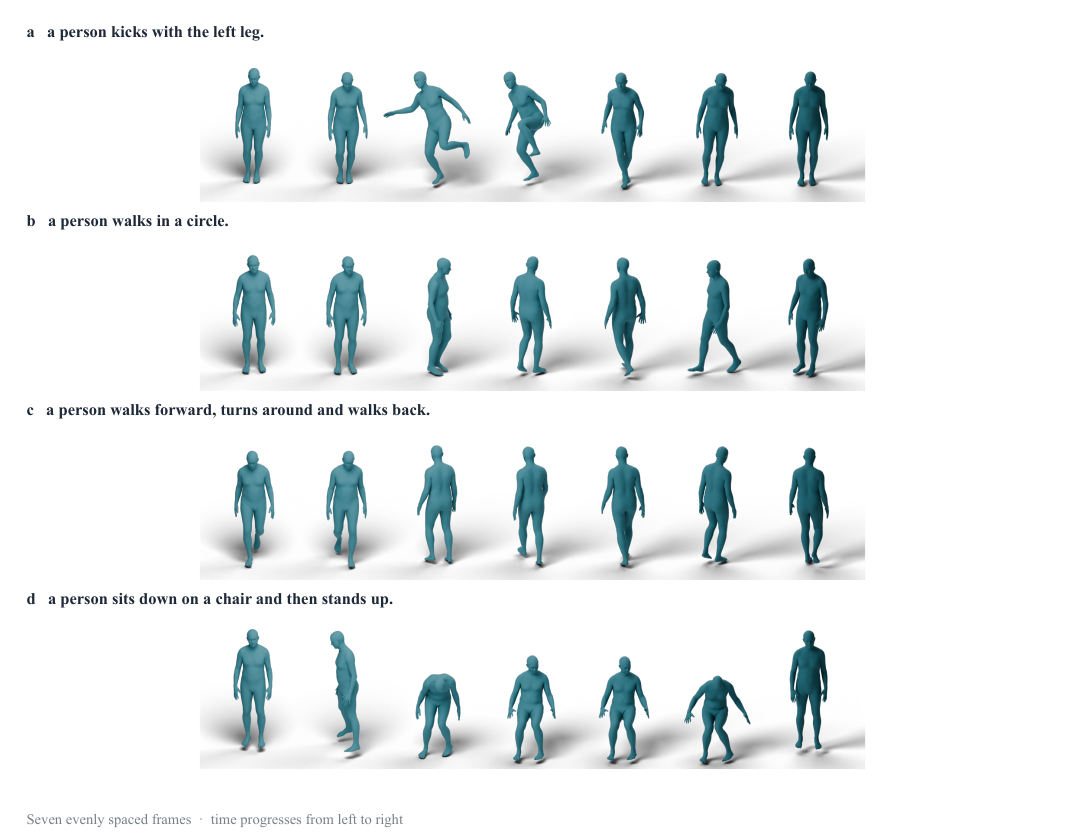}
\caption{\textbf{Limb actions and motion transitions.} Four MoRAE-32 examples, five RK4 steps, guidance 4.5. The first sample per prompt is shown at seven evenly spaced frames; time progresses from left to right.}
\label{fig:supp_qual_more}
\end{figure*}

\begin{figure*}[!htbp]
\centering
\includegraphics[width=\textwidth]{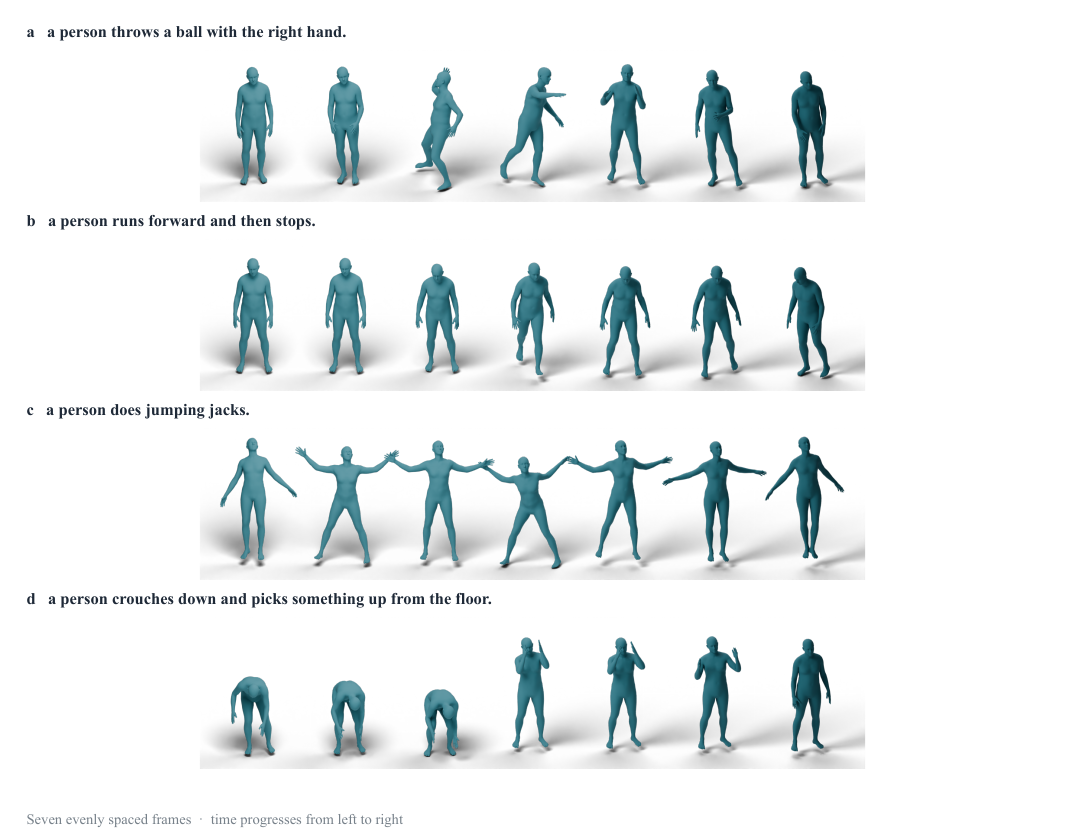}
\caption{\textbf{Whole-body actions and coordinated movement.} Four additional MoRAE-32 examples under the same generation and visualization settings as \cref{fig:supp_qual_more}.}
\label{fig:supp_qual}
\end{figure*}

\Cref{fig:supp_qual_more,fig:supp_qual} show eight MoRAE-32 examples at five RK4 steps and guidance 4.5, each displaying the first sample at seven evenly spaced frames.

\paragraph{Surface fitting.}
For mesh visualization, we fit a neutral SMPL body surface to the generated 22-joint trajectories using a multi-stage SMPLify3D solver.
It first optimizes torso translation and global orientation, followed by 30 outer L-BFGS steps (up to 30 iterations each) with GMM pose, knee/elbow angle, shape, and pose-preservation priors.
The first-frame body shape is fixed across the sequence, and each frame initializes from the previous one.
Mesh fitting is used purely for visualization; all quantitative evaluations operate directly on the generated joint features.

\paragraph{Main-paper qualitative comparison.}
The prompt in \cref{fig:qual} is one of 50 prompts that we wrote to cover ten kinds of difficulty (five prompts each); every prompt has a character similarity below 0.85 to all HumanML3D captions.
Each method generates a single 196-frame sample per prompt with batch size one and the same random seed, and no sample is re-drawn.
The five prior methods use their official code, released checkpoints, and documented sampling settings; \ours uses the 64-D model with five RK4 steps and guidance $w{=}4.5$.
No method receives foot IK or any other post-processing.
The figure shows frames 0, 28, 56, 84, 111, 139, 167, and 195; the meshes of all six methods come from the same batched SMPL fitting with a temporal-consistency term.

\begin{figure*}[!htbp]
\centering
\includegraphics[width=\textwidth]{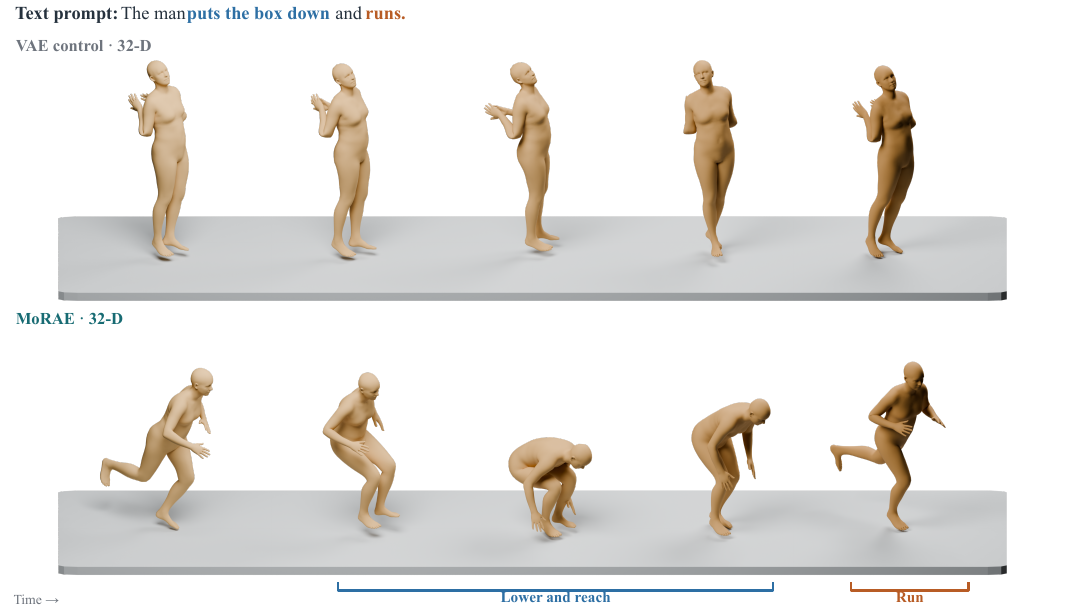}
\caption{\textbf{Comparison against a VAE latent with the generator matched.} For ``the man puts the box down and runs,'' \ours bends down to place the box before running, whereas the VAE sample stays upright. Both models use 32-D latents and share the random seed, initial noise, and sampler.}
\label{fig:supp_qual_vae}
\end{figure*}

\paragraph{Matched comparison against a reconstruction latent.}
\Cref{fig:supp_qual_vae} holds the generator fixed and changes only the latent.
Both 32-D models generate a 64-frame motion for the same prompt from identical initial noise (seed 11) with a matched DiT training seed, five RK4 steps, guidance $w{=}4.5$, and neutral body shape, rendered at frames 0, 10, 21, 40, and 63.
This example was selected from a candidate pool of 66 test prompts (three noise seeds each) and 12 multi-action prompts.
Rendered meshes are horizontally offset for display while preserving global orientation and relative joint positions.

% ===== end sec/X_suppl.tex =====

\end{document}